\documentclass[twoside]{article}

\usepackage[accepted]{aistats2022}

\usepackage{tikz}
\usetikzlibrary{positioning,chains,shapes.misc,fit}
\usepackage[utf8]{inputenc} 
\usepackage[T1]{fontenc}    
\usepackage{hyperref}       
\usepackage{url}            
\usepackage{booktabs}       
\usepackage{amsfonts}       
\usepackage{nicefrac}       
\usepackage{microtype}      
\usepackage{xcolor}         
\usepackage{xspace}
\usepackage{amsmath}
\usepackage{amsthm}
\usepackage{multirow}
\newcommand*{\eg}{e.g.\@\xspace}
\newcommand*{\ie}{i.e.\@\xspace}
\usepackage{graphicx}
\usepackage{caption}
\usepackage{subcaption}
\usepackage[inline]{enumitem}

\theoremstyle{remark}
\newtheorem{remark}{Remark}[subsection]

\usepackage[round]{natbib}

\begin{document}

%

%

\twocolumn[

\aistatstitle{It's FLAN time! Summing feature-wise latent representations for interpretability}

\aistatsauthor{ An-phi Nguyen \And Stefania Vasilaki \And  Maria Rodr\'iguez Mart\'inez}

\aistatsaddress{ IBM Research Europe, ETH Z\"urich \And  ETH Z\"urich \And IBM Research Europe } ]

\begin{abstract}
Interpretability has become a necessary feature for machine learning models deployed in critical scenarios, \eg legal system, healthcare. In these situations, algorithmic decisions may have (potentially negative) long-lasting effects on the end-user affected by the decision.
In many cases, the representational power of deep learning models is not needed, therefore \emph{simple and interpretable} models (\eg linear models) should be preferred. However, in high-dimensional and/or complex domains (\eg computer vision), the universal approximation capabilities of neural networks are required. Inspired by linear models and the Kolmogorov-Arnold representation theorem, we propose a novel class of \emph{structurally-constrained} neural networks, which we call FLANs (Feature-wise Latent Additive Networks). Crucially, FLANs process each input feature \emph{separately}, computing for each of them a representation in a common latent space. These feature-wise latent representations are then simply \emph{summed}, and the aggregated representation is used for prediction. These constraints (which are at the core of the interpretability of linear models) allow a user to estimate the effect of each individual feature \emph{independently} from the others, \emph{enhancing interpretability}. In a set of experiments across different domains, we show how without compromising excessively the test performance, the structural constraints proposed in FLANs indeed facilitates the interpretability of deep learning models. We quantitatively compare FLANs interpretability to post-hoc methods using recently introduced metrics, discussing the advantages of natively interpretable models over a post-hoc analysis.
\end{abstract}

\section{INTRODUCTION}
The recent surge in interest towards \emph{interpretable machine learning} research can arguably be attributed to the success of deep learning models. Despite their universal approximation capabilities \citep{Cybenko1989ApproximationFunction,Hornik1989MultilayerApproximators} and generalization properties \citep{Kawaguchi2017GeneralizationLearning,Arora2018StrongerApproach}, these models often behave as a black-box from a user perspective: their ``decision process'' is often unclear, especially to layman users. While not always necessary, interpretability can be \emph{critical} in applications with far-reaching impact, \eg in legal cases and healthcare. In these high-stake scenarios, \citep{Rudin2019StopInstead} advocates for \emph{simple} and \emph{already interpretable} models. Indeed, for various real-world problems, simple interpretable models perform similarly to more complex models \citep{Semenova2019ALearning} and should therefore be preferred. 

Unfortunately, this solution is not applicable for high-dimensional and/or complex problems where the representational capabilities of end-to-end deep learning models give them an edge against other models, \eg autonomous driving \citep{Gupta2021DeepIssues} or machine translation \citep{Wu2016GooglesTranslation}. In these cases, interpretability can be achieved in two ways. \emph{Post-hoc} methods, such as feature attribution methods \citep{Ancona2018TowardsNetworks}, can be leveraged to produce an explanation for a model that \emph{has already been trained}. Alternatively, it is possible to directly train complex \emph{ante-hoc} models, which are models with some kind of ``in-built mechanism'' that allows for their scrutiny. Classical examples are decision trees \citep{Quinlan1986InductionTrees}. Recently, there has been an increased interest in developing ante-hoc deep learning models that attempt at retaining the representational and generalization properties of neural networks, whilst still being interpretable \citep[e.g.][]{Chen2019ThisRecognition,AlvarezMelis2018TowardsNetworks,Nguyen2019MonoNet:Features}.

In this paper, we introduce a model, which we denominate \emph{FLAN} (Feature-wise Latent Additive Network), that belongs to this last class of interpretable models. Inspired by linear models, FLAN computes a latent representation of each feature (or subsets thereof) \emph{separately}. The latent representation of an input sample is obtained by simply \emph{summing} the feature representations, and then passed to a classifier network for classification. We posit that these two constraints are what enables the interpretability of our model (Section~\ref{sec:linear_motivation}). Despite these constraints, our model can still achieve good performance on a series of benchmark tasks in different domains. 

\section{FLANs}\label{sec:model}

\begin{figure}
\resizebox{\linewidth}{!}{
     \centering
     \begin{tikzpicture}[shorten >=1pt,->, draw=black!50, 
        node distance = 12mm and 17mm,
          start chain = going below,
every pin edge/.style = {<-,shorten <=1pt},
        neuron/.style = {circle,font=\large,
                         minimum size=17pt, 
                         on chain},
        operation/.style = {circle,font=\large,draw=black!100,
                         minimum size=7pt, inner sep=0pt,
                         on chain},
        function/.style = {rounded corners=.05cm, fill=#1,font=\Large,
                         minimum size=23pt, inner sep=0pt,
                         on chain},
        sample/.style = {minimum size=17pt,font=\Large, 
                         on chain},
         annot/.style = {text width=4em, align=center}
                        ]
    \node[neuron] (I-1)    {$x_{1}$};
    \node[neuron] (I-i)    {$x_{i}$};
    \node[neuron] (I-N)    {$x_{N}$};
    \path (I-1) -- (I-i) node [black, midway, sloped] {$\dots$};
    \path (I-i) -- (I-N) node [black, midway, sloped] {$\dots$};
    
    \node[sample,left=of I-i] (x) {$\mathbf{x}$};
    
    \node[function=black!10,right=of I-1] (ph-1) {$\phi_1$};
    \node[function=black!10,right=of I-i] (ph-i) {$\phi_i$};
    \node[function=black!10,right=of I-N] (ph-N) {$\phi_N$};
    \path (ph-1) -- (ph-i) node [black, midway, sloped] {$\dots$};
    \path (ph-i) -- (ph-N) node [black, midway, sloped] {$\dots$};
    
    \node[operation,right=of ph-i] (sum) {$+$};
    
    \node[function=black!10,right=of sum] (psi) {$\psi$};
    
    \coordinate[right of=psi] (out);
    \path (x) edge (I-1);
    \path (x) edge (I-i);
    \path (x) edge (I-N);
    \path (I-1) edge (ph-1);
    \path (I-i) edge (ph-i);
    \path (I-N) edge (ph-N);
    \path (ph-1) edge (sum);
    \path (ph-i) edge (sum);
    \path (ph-N) edge (sum);
    \path (sum) edge (psi);
    
    \path (psi) edge (out);
    \end{tikzpicture}
    }
     
     \caption{The Base Architecture of FLANs. A sample $\mathbf{x}$ is split in its features $x_i$. Each of the feature is fed to a different function $\phi_i$. All the processed features are then aggregated by summation, and finally fed to a prediction network $\psi$. In our work, all the functions $\phi_i$ and $\psi$ are implemented as neural networks.}
     \label{fig:base_flan}
     
\end{figure}
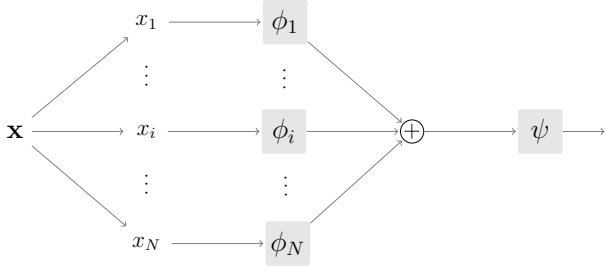

\subsection{Interpretability Lessons from Linear Models}\label{sec:linear_motivation}

Linear models are generally considered among the prime examples of interpretable models. Arguably this is due to two key characteristics of these models:

\begin{itemize}
    \item \emph{Separability of features:} Linear models do not take \emph{interactions} among features into account (unless an interaction term is explicitly added to the features). This means that a user can examine the effect of each feature \emph{separately} from the others. This, in turn, makes the model easier to understand from a user perspective \citep{Schulz2017CompositionalLearning}
    \item \emph{Predictability of the output:}  Given a single feature, its effect on the output can be easily understood by a human user if the relationship is linear \citep{Byun1995InteractionLearning}. In the presence of multiple features, the human can still easily predict their effect on the output(s), if their effect is separable as discussed in the previous point. 
\end{itemize}

\subsection{Model Architecture}\label{sec:architecture}

Motivated by the previous section, we want to build a model for which the effect of single features can be predicted separately from each other. Let us assume that our goal is to learn a function $f: \mathcal{X} \rightarrow \mathcal{Y}$, where $\mathcal{X}$ has dimension $N$ and $\mathcal{Y}$ has dimension $M$. We propose to implement the function $f$ as:
\begin{equation}\label{eq:lanet}
    f(\mathbf{x}) = f(x_1, \dots, x_i, \dots, x_N) = \psi \Big( \sum_{i=1}^N  \phi_i (x_i)\Big)
\end{equation}

where:

\begin{itemize}
    \item $x_i$ $(\mathrm{with} \; i=1,\dots,N)$ are the features, \ie the components of the input sample $\mathbf{x}$;
    \item $\phi_i: \mathcal{X}_i \rightarrow \mathcal{Z}$ are feature functions that act on each individual feature $x_i$ separately and map them to the \emph{same} latent space $\mathcal{Z}$ of dimension $D$. Note that the feature functions may either implement different functions, or the same function for all features;
    \item the aggregate $\sum_{i=1}^N \phi_i (x_i) = \mathbf{z} \in \mathcal{Z}$ is the latent representation for the input sample $\mathbf{x} $;
    \item the predictor function $\psi: \mathcal{Z} \rightarrow \mathcal{Y}$ maps the sample latent representation to the output space $\mathcal{Y}$. 
\end{itemize}

In this paper, we implement the feature functions $\phi_i$ and the prediction function $\psi$ as neural networks and learn them in an end-to-end fashion. Figure~\ref{fig:base_flan} provides a depiction of our model.

Before discussing the interpretability of our model (Section~\ref{sec:interpreting_sum}), we would like to make a few remarks.

\begin{remark}[\textbf{Universal Approximation}]~\label{remark:approx} Since no interaction is explicitly modeled in Eq.~\eqref{eq:lanet}, a concern that may be raised is if our model still retains the same approximation capabilities of traditional neural networks. The answer is given by the \emph{Kolmogorov-Arnold representation theorem} \citep{Kolmogorov1957OnAddition} which claims, informally, that any continuous function of a finite number of variables can be expressed in the form 
\begin{equation*}
    f(\mathbf{x}) = f(x_1, \dots, x_N) = \sum_{q=0}^{2N} \Phi_q \Big(\sum_{i=1}^N \phi_{q,i} (x_i)\Big)
\end{equation*}
Note that for interpretability purposes, we do not need the outer sum since it would not separate neither the outputs nor the effects of the input features. We therefore use a more generic function $\psi$. The strength of the Kolmogorov-Arnold representation theorem is in claiming that there is actually \emph{no need} to explicitly model interactions. Although there is some debate about the applicability of the Kolmogorov-Arnold theorem \citep[e.g.][]{Girosi1989RepresentationIrrelevant,Kurkova1991KolmogorovsRelevant}, in Section~\ref{sec:experiments} we experimentally show that our model can (over)fit the training dataset.

\end{remark}

\begin{remark}[\textbf{Feature Subgroups}]~\label{remark:feat_groups} In some applications, having a function applied to each individual feature may be detrimental. From an interpretability perspective, this is especially true when a single feature has no particular meaning. For example, consider computer vision tasks: human users rarely understand image classification in terms of single pixels, but rather in terms of higher-level concepts. Eq.~\eqref{eq:lanet} can be adapted to consider features in \emph{(non-overlapping) groups} rather than individually. Building on the computer vision example, an image could be processed in \emph{patches} rather than single pixels. 
\end{remark}

\begin{remark}[\textbf{Structural Information}]~\label{remark:structure} Some data types carry additional structural information, \eg natural language or images. Eq.~\eqref{eq:lanet} can be applied directly also to these domains. However, for the purpose of parameter sharing, it may be useful to make the dependency on the structure explicit:
\begin{equation*}
    f(\mathbf{x}) = f(x_1, \dots, x_N) = \psi \Big( \sum_{i=1}^N  \phi (x_i; \theta, p_i)\Big)
\end{equation*}
where $\theta$ are parameters common to all the feature functions, and $p_i$ encode some kind of structural information. This is the same idea behind the positional embeddings used in Transformer architectures \citep{Vaswani2017AttentionNeed}.

\end{remark}



\subsection{Interpreting FLANs}\label{sec:interpreting_sum}

In this section, we will discuss the three main modalities to interpret FLANs. Concrete examples will then be presented in Section~\ref{sec:experiments}.

\subsubsection{Separating and Predicting the Affect of \emph{Individual} Features}\label{sec:interpret_mech}

The interpretability of FLANs stems from the fact that different features are processed \emph{separately}. That is, we can analyze how each feature contributes to the prediction without being concerned about interactions. 

A user can study the effect of a single feature $x_i$ simply by performing a prediction on that feature, \ie $\psi(\mathbf{z}_i)$ with $\mathbf{z}_i = \phi_i(x_i)$. Note however that, since we are not making any assumption on the function $\psi$, the effect is not generally additive. In fact, assume that $\mathcal{Y}$ and $\mathcal{Z}$ are equipped, respectively, with norms $||\cdot||_{\mathcal{Y}}$ and $||\cdot||_{\mathcal{Z}}$. Further, let us informally assume that we can compute the Taylor expansion of $\psi$ in a ``large enough'' neighborhood of $\mathbf{z}_{*} = \sum_{j=1, \; j\neq i}^N \mathbf{z}_j$. Then we have
\begin{equation}\label{eq:approx_interpret}
\begin{split}
    ||\underbrace{\psi(\mathbf{z}_{*} + \mathbf{z}_i) - \psi(\mathbf{z}_{*})}_{(\Delta)} - \psi(\mathbf{z}_i)||_{\mathcal{Y}} & = \\ ||\mathbf{J}_{\mathbf{z}_*}\mathbf{z}_i - \psi(\mathbf{z}_i) + o(||\mathbf{z}_{*}||_{\mathcal{Z}})||_{\mathcal{Y}}
\end{split}   
\end{equation}
where:
\begin{itemize}
    \item $\psi(\mathbf{z}_{*} + \mathbf{z}_i)- \psi(\mathbf{z}_{*})$ is the change in prediction given by the additional information contained in the $i$-th feature;
    \item $\mathbf{J}_{\mathbf{z}_*}$ is the Jacobian of $\psi$ computed at point $\mathbf{z}_{*}$;
    \item $o(||\mathbf{z}_{*}||_{\mathcal{Z}})$ is the remainder term in the first-order Taylor expansion.
\end{itemize}
The right-hand side of Eq.~\eqref{eq:approx_interpret} gets closer to zero the ``more linear''  is $\psi$. Therefore Eq.~\eqref{eq:approx_interpret} is telling us that we can estimate the change in prediction $(\Delta)$ given by feature $i$ by looking at the prediction $\psi(\mathbf{z}_i)$ on that feature. The accuracy of the estimation will depend on how non-linear is $\psi$ . Although, we are not making any assumption on $\psi$, in Section~\ref{sec:interpret_results} we show that in practice we can still estimate the effect of single features on the prediction.

\subsubsection{Feature Importance}\label{sec:feature_imp}

The interpretability modality described in the previous section has mostly an ``algorithmic/mechanistic'' flavor. However, FLANs can be interpreted also by computing feature importances without the need for \emph{post-hoc} methods, such as SHAP \citep{Lundberg2017APredictions}. Since sample representations are simply the \emph{sum} of feature representations, then a feature $i$ that is mapped to a small vector $||\mathbf{z}_i||_{\mathcal{Z}}\approx 0$ brings no contribution to the prediction. This means that we can use the norms of the feature latent representations as indicative of feature importance.

\subsubsection{Example-Based}\label{sec:example_based}

FLANs can be further interpreted via examples/prototypes: to interpret the model at an input sample $\mathbf{\hat{z}}$, we only need to look for the nearest samples $\mathbf{\tilde{z}}$ in the latent space. Since $\psi$ in our model is a neural network, and therefore a Lipschitz continuous function, \ie $||\psi(\mathbf{\hat{z}}) - \psi(\mathbf{\tilde{z}})||_\mathcal{Y} \leq L ||\mathbf{\hat{z}} - \mathbf{\tilde{z}}||_\mathcal{Z}$, the predictions performed on two samples with similar representations will be similar.

This analysis can be performed also on a feature level. That is, by looking for features with similar representations, we can understand which features provide similar information towards the prediction. 
Similarly, by looking at \emph{dissimilar} features, we are able to understand why two samples may be classified differently.

\section{RELATED WORK}\label{sec:rel_work}

Our work is motivated by interpretability. In the taxonomy \citep{Molnar2019InterpretableLearning} of interpretability methods, FLANs classifies primarily as a \emph{local} method since the model can be explained at single sample points. A more \emph{global} overview of the model can be obtained by looking for similar samples in the latent space, as explained in Section~\ref{sec:example_based}. Here, any of previously proposed methods can be used (\eg MMD-Critic by \citealt{Kim2016ExamplesInterpretability}, ProtoDash by \citealt{Gurumoorthy2017EfficientWeights}). 

FLANs can be inspected also via feature attributions (Section~\ref{sec:feature_imp}). These feature importances are computed as norms of the feature representations. Therefore there is no need to use any post-hoc method, such as gradient-based methods \citep{Ancona2018TowardsNetworks,Selvaraju2017Grad-CAM:Localization,Bhatt2019TowardsAttributions,Sundararajan2017AxiomaticNetworks} or SHAP \citep{Lundberg2017APredictions}, which would require more computational resources. This advantage is shared with Transformer models \citep{Vaswani2017AttentionNeed}, where the attention scores can be interpreted as importances. However, the attention scores are function of \emph{all} the feature, effectively modeling interactions. This, in turn, decreases separability and, consequently, interpretability.

From a modeling perspective, FLANs belong to the class of \emph{ante-hoc structurally constrained} deep learning models. Notable methods in this category are those that augment the network with prototype/prototypical-part based reasoning, \eg \citep{Li2017DeepPredictions,Chen2019ThisRecognition}. A similar mechanism can be achieved also with FLANs (Section~\ref{sec:example_based}) without the need of an \emph{ad-hoc} cost function or training strategy. Prototypes are similar in spirit to the concepts/feature basis used in Self-Explaining Neural Networks (SENNs) \citep{AlvarezMelis2018TowardsNetworks}. Interestingly, \citep{AlvarezMelis2018TowardsNetworks} also recognize the value of additivity/separability for interpretability. However, additivity is enforced only in the last layer in their model. The concepts (and their relevances) are computed as functions of the entire input space. Consequently, the interpretability with respect to the original input space is lost. 


Closer in spirit to our model are the neural additive models \citep{Agarwal2020NeuralNets}, Explainable Boosting Machines (EBM) \citep{Nori2019InterpretML:Interpretability}, and the generalized additive models with interactions \citep{Lou2013AccurateInteractions}. 
However, these previous works do not leverage a second prediction function after aggregating the per-feature functions, thus potentially losing the approximation capabilities of FLANs. Moreover, the authors demonstrate their model only on tabular datasets, while we show that additivity-based models can be successfully used also in other more complex domains.


\section{EXPERIMENTAL RESULTS}\label{sec:experiments}

\subsection{FLAN Benchmarking Results}\label{sec:perf_results}

\paragraph{Datasets.} We test FLANs on a series of benchmark datasets across different domains. In the tabular domain, we consider ProPublica's Recidivism Risk score prediction dataset (\texttt{COMPAS})\footnote{\url{https://github.com/propublica/compas-analysis/}}, and different risk datasets from the UCI benchmark repository \citep{Dua2017UCIRepository} (\texttt{heart},\texttt{adult},\texttt{mammo}). For image classification, we benchmark FLANs on the two standard digit classification datasets \texttt{MNIST} \citep{LeCun1998Gradient-basedRecognition} and \texttt{SVHN} \citep{Netzer2011ReadingLearning}, and the fine-grained bird recognition task \texttt{CUB-200-2011} \citep{Welinder2010Caltech-UCSD200}. For text datasets, we consider \texttt{AGNews} \citep{Zhang2015Character-LevelClassification} and \texttt{IMDb} \citep{Maas2011LearningAnalysis}. For the textual domain, we further consider the healthcare-oriented \texttt{TCR-Epitope} dataset, where the task is to predict two amino-acid sequences will bind (binary classification). We refer to \citet{Weber2021TITAN:Networks} for more details. The results are summarised in Table~\ref{tab:benchmark}. 

\paragraph{Models for Comparison and Training.} We compare FLANs to both established and state-of-the-art methods. The chosen models for each domain are reported in Table~\ref{tab:benchmark}. For tabular datasets, we train all the models ourselves. For the other tasks, we only train FLANs and retrieve the performance of other models from the literature. Training details (Appendix~\ref{app:train_det}) and further results (Appendix~\ref{app:more_res}) are provided in the appendix. However, it is important to specify how FLANs splits the features in each task. For tabular tasks, FLANs simply process each feature individually. Similarly, in text datasets, our model considers each token (\ie either a word for \texttt{AGNews} and \texttt{IMDb}, or a single amino-acid for \texttt{TCR-Epitope}) in the sentence/sequence as a single feature. For image datasets, \emph{non-overlapping} square patches are the features fed to FLANs (Remark~\ref{remark:feat_groups}). 

\paragraph{Tabular Datasets Results.} For benchmarking on tabular datasets, we follow \citet{Agarwal2020NeuralNets} and measure the performance of different models in terms of Area Under the Curve (AUC). Table~\ref{tab:tab_benchmark} shows that these datasets are easy enough that a simple logistic regression model can perform well. The results on the \texttt{adult} and \texttt{mammo} datasets suggest that linearity is a good inductive bias for these tasks since logistic regression is able to consistently outperform all the other (non-linear) models. On the other hand, in the \texttt{heart} dataset and, to a lesser degree, in the \texttt{COMPAS} dataset, it seems beneficial to include non-linearities and interactions. In particular, FLANs closely replicates the performance of more traditional feedforward networks (MLP). This might suggest that FLANs are similar to MLPs in terms of approximation capabilities.

\paragraph{Image Datasets Results.} FLANs results (Table~\ref{tab:img_benchmark}) on the \texttt{MNIST} dataset are comparable to established methods. Moreover, linear models (results not reported) do not achieve more than 94\% test accuracy, providing further evidence to the ability of FLANs in implementing interactions \emph{without explicitly modeling them}. We further tested our model on the more difficult fine-grained image classification dataset \texttt{CUB-200-2011}. FLANs do not achieve the same accuracy as other models. This might be explained by the fact that the models reported are pretrained on \texttt{ImageNet} \citep{Russakovsky2014ImageNetChallenge} and further fine-tuned on this dataset. On the other hand, in our experiments, our top-performing models use only \emph{some layers} of a pretrained ResNeXt \citep{Xie2017AggregatedNetworks} as part of the patch feature function $\phi_i$. We hypothesize that the inferior performance of FLANs is attributed to the fact that our model has to essentially \emph{learn the interactions from scratch}, and \texttt{CUB-200-2011} might be a too small of a dataset to effectively learn this. Despite the lower accuracy, and given the relatively small size of the dataset (11.7k images split across 200 classes), we see our results as promising and a good basis for future investigations in \emph{large scale} image recognition tasks that require interpretability.

\paragraph{Text Datasets Results.} On the considered benchmark datasets, FLANs fair well against traditional LSTM/CNN-based models. The drop in performance of FLANs is particularly noticeable on the \texttt{IMDb} against more modern attention-based architectures. Considering that \texttt{IMDb} contains much longer sentences than \texttt{AGNews}, these results suggest that FLANs may have difficulties in learning longer term dependencies/interactions. This is consistent with the conclusion reached on image datasets.

\paragraph{TCR-Epitope Dataset Results.} FLAN outperforms the K-Nearest Neighbor baselines by a large margin and achieves comparable results to TITAN~\citep{Weber2021TITAN:Networks}, the state-of-the-art deep model for the task, \emph{even without pretraining or augmentation strategies}. 

\begin{table}[h]
  \caption{Benchmarking Results. (*) denotes pretrained models or models using automated augmentation strategies \citep[e.g.][]{Cubuk2020RandAugment:Space}. (**) denotes FLAN models that use pretrained models as \emph{part} of the feature functions.}\label{tab:benchmark}
  
  \begin{subtable}{\linewidth}
  \centering
  \caption{Area Under The Curve (AUC) on Tabular Datasets.}\label{tab:tab_benchmark}
  
  \resizebox{\linewidth}{!}{
  \begin{tabular}{lcccc}
    \toprule
         & \texttt{COMPAS}  & \texttt{adult} & \texttt{heart} & \texttt{mammo} \\
    \midrule
    Logistic Regression
    & 0.905 & 0.892 & 0.873 & \textbf{0.841} \\
    Decision Tree (small)
    & 0.903 & 0.865 & 0.849 & 0.799 \\
    Decision Tree (unrestricted)
    & 0.902 & 0.813 & 0.848 & 0.801 \\
    Random Forest
    & \textbf{0.915} & 0.869 & 0.945 & 0.822 \\
    EBM~\citep{Nori2019InterpretML:Interpretability}  & 0.911 & \textbf{0.893} & 0.941 & 0.840  \\
    MLP  & \textbf{0.915} & 0.874 & 0.937 &  0.831 \\
    SENN~\citep{AlvarezMelis2018TowardsNetworks}  & 0.910 & 0.865 & 0.881 &  0.834 \\
    \midrule
    FLAN  & 0.914 & 0.880 & \textbf{0.950} &  0.832 \\
    \bottomrule
  \end{tabular}
  }
  \end{subtable}
  
  \vspace{1ex}
  
  \begin{subtable}{\linewidth}
  \centering
  \caption{Test Accuracy (\%) on Image Datasets.}\label{tab:img_benchmark}
  \resizebox{\linewidth}{!}{
  \begin{tabular}{lccc}
    \toprule
         & \texttt{MNIST} & \texttt{SVHN} & \texttt{CUB} \\
    \midrule
    ResNet~\citep{He2016DeepRecognition,Wang2020ICapsNets:Classification} & 99.2 & 94.5* & 84.5* \\ 
    iCaps~\citep{Wang2020ICapsNets:Classification}  & 99.2 & 92.0 & - \\
    ViT~\citep{Dosovitskiy2021ANSCALE,He2021TransFG:Recognition} & - & 88.9 & 90.4* \\
    ProtoPNet~\citep{Chen2019ThisRecognition} & - & - & 84.8*\\
    SENN~\citep{AlvarezMelis2018TowardsNetworks} &  99.1 &  - & - \\
    SotA~\citep{Byerly2020ACapsules,Cubuk2020RandAugment:Space} & \textbf{99.84} & \textbf{99.0}* & \textbf{91.3}*\\
    \midrule
    FLAN &  99.05 & 93.41 & 71.53** \\
    \bottomrule
  \end{tabular}
  }
  \end{subtable}
  
  \vspace{1ex}
  
  \begin{subtable}{\linewidth}
  \centering
  \caption{Test Accuracy (\%) on Text Datasets.}\label{tab:text_benchmark}
  \resizebox{\linewidth}{!}{
  \begin{tabular}{lcc}
    \toprule
         & \texttt{AGNews} & \texttt{IMDb} \\
    \midrule
    CharCNN \citep{Zhang2015Character-LevelClassification} & 90.49 & - \\
    LSTM \citep{
    Wang2018DisconnectedCategorization,Dai2015Semi-supervisedLearning} & 93.8 & 86.5 \\
    VDCNN \citep{Conneau2017VeryClassification,Abreu2019HierarchicalClassification} & 91.33 & 79.47 \\
    HAHNN \citep{Abreu2019HierarchicalClassification} & - & 95.17 \\
    XLNet \citep{Yang2019XLNet:Understanding} & \textbf{95.6}* & \textbf{96.8}* \\
    \midrule
    FLAN  & 91.2 &  85.2 \\
    \bottomrule
  \end{tabular}
  }
  \end{subtable}
  
  \vspace{1ex}

  \begin{subtable}{\linewidth}
  \centering
  \caption{AUC on the \texttt{TCR-Epitope} Dataset.}\label{tab:tcr_benchmark}
  
  \resizebox{\linewidth}{!}{
  \begin{tabular}{lcc}
    \toprule
         & \texttt{TCR-Epitope}  
         \\
    \midrule
    KNN ($K=7$) & 0.770 
    \\
    KNN ($K=21$)& 0.779 
    \\
    TITAN \cite{Weber2021TITAN:Networks} & 0.841 
    \\
    Augmented TITAN \citep{Weber2021TITAN:Networks}  & \textbf{0.867*} 
    \\
    \midrule
    FLAN  & 0.858 
    \\
    \bottomrule
  \end{tabular}
  }
  \end{subtable}
  
\end{table}

\paragraph{Summary.} FLANs managed to achieve results comparable to the most established models on a series of benchmark datasets. Results against more recent architectures suggest that FLANs may have difficulties in learning complex/long interactions. However, it is worth noting that FLANs do manage to achieve 100\% training accuracy in our experiments (Appendix~\ref{app:more_res}). This suggests that our model is able to learn complex interactions, although they may not be \emph{generalizable}. 

\subsection{Interpretability Results}\label{sec:interpret_results}

\subsubsection{COMPAS}

\begin{figure}[h]
     \centering
     \begin{subfigure}[b]{\linewidth}
         
         \centering
         \includegraphics[width=\linewidth]{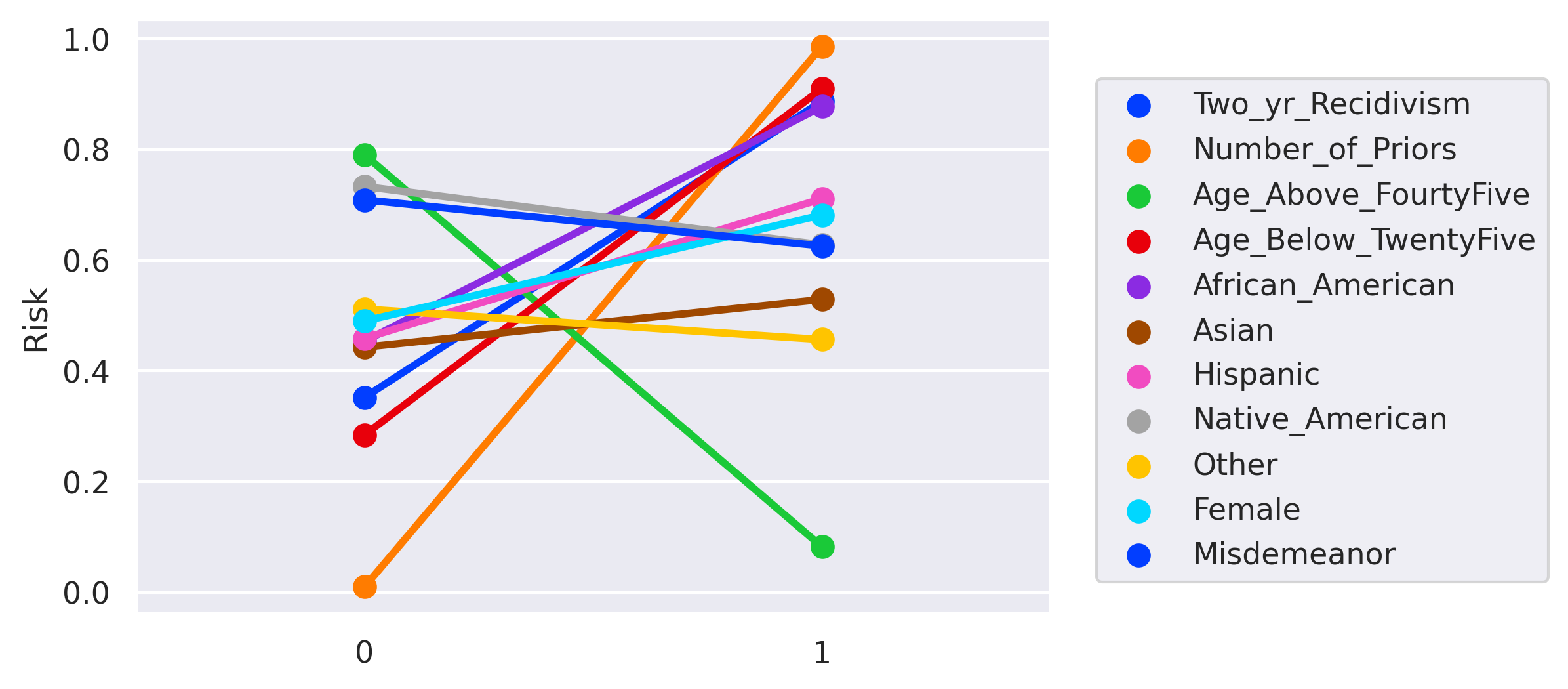}
         \caption{Feature Effects}
         \label{fig:compas_eff}
         
     \end{subfigure}
     \hfill
     \begin{subfigure}[b]{\linewidth}
         
         \centering
         \includegraphics[trim=0 -0.6cm 0 0,clip,width=0.8\linewidth]{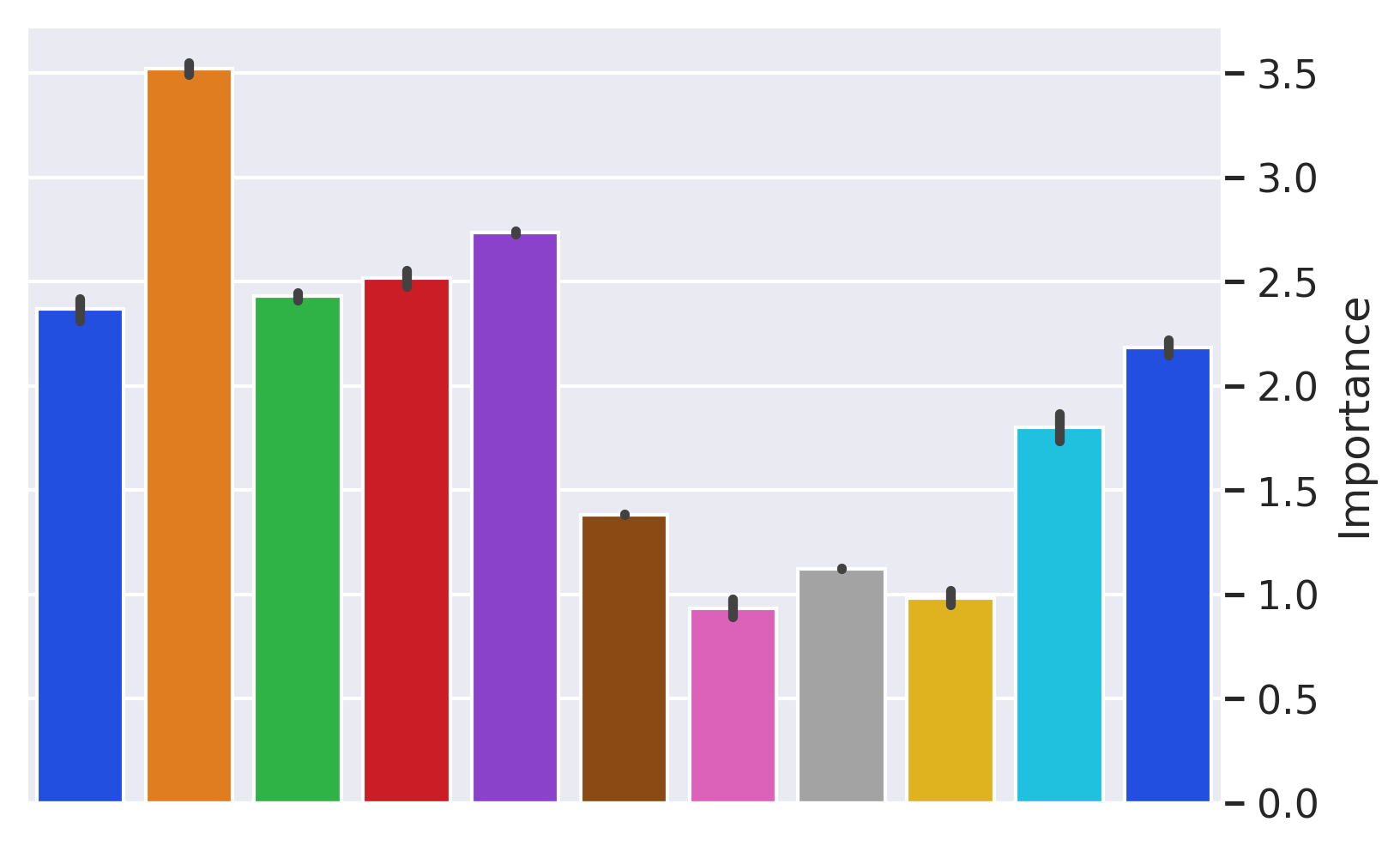}
         \caption{Feature Importances}
         \label{fig:compas_imp}
         
     \end{subfigure}
     
         \caption{Interpreting A FLAN Model Trained On The COMPAS Dataset. (a) Effect of the single features. (b) Feature importances averaged over the training set. The legend in the center shows the colors used to denote each feature.}
    \label{fig:compas}
    
\end{figure}

We use the \texttt{COMPAS} dataset as a propedeutic example to show how to interpret FLANs. As discussed in Section~\ref{sec:interpret_mech}, we can study the approximate effect of single features separately, by applying the prediction network $\psi$ to the feature latent representation. In this case study, this is even easier since all the features are binarized. Figura~\ref{fig:compas_eff} shows how the predicted risk changes if we switch a feature from 0 to 1. The results suggest that the risk is \emph{increased} for criminals that have a high number of priors, are younger than 25, are Afro-American, or have already re-offended in the past two years. Interestingly, the risk seems particularly decreased for criminals above the age of 45. To further validate these findings we analyze the feature importances provided by our model. For each sample, we compute the importances as explained in Section~\ref{sec:feature_imp} and then we average them over the training set. Figure~\ref{fig:compas_imp} confirms that across the training set, the features mentioned above are the most discriminative ones. Previous analyses performed using interpretable models \citep{Agarwal2020NeuralNets,AlvarezMelis2018TowardsNetworks} reached similar conclusions.

\subsubsection{CUB}\label{sec:exp_cub}

We qualitatively validate the interpretability capabilities of our model on the more complex \texttt{CUB-200-2011} dataset. For further qualitative results, please refer to Appendix~\ref{app:more_res}.

\begin{figure}[h]
    \centering
    \includegraphics[width=\linewidth]{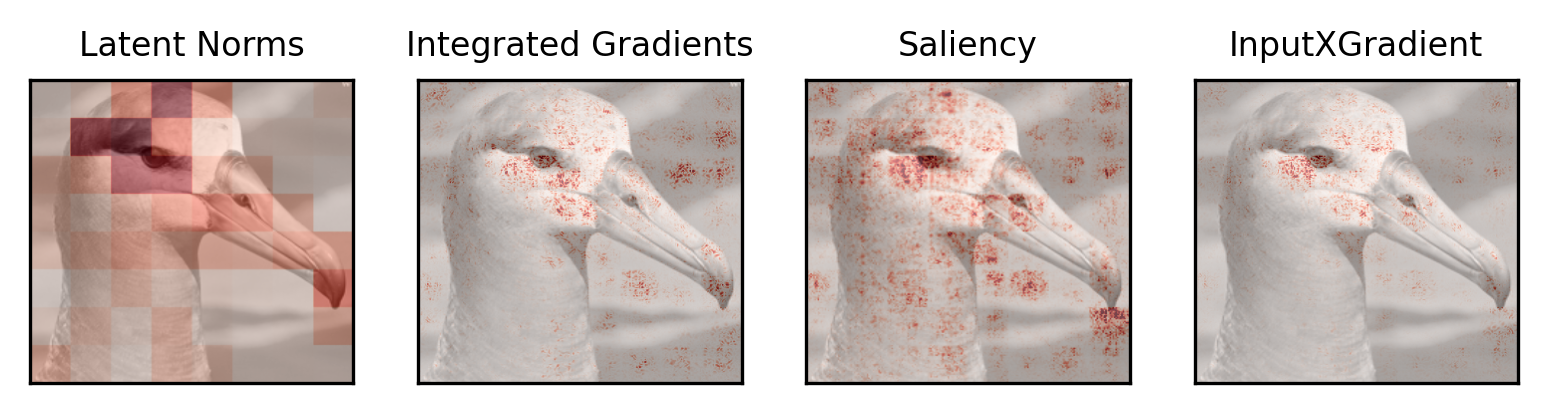}
    
    \caption{Comparison of the feature importances computed by our model (Latent Norms) against 3 established gradient-based feature attribution methods.}
    \label{fig:cub_attr_comparison}
\end{figure}

\paragraph{Feature Importances.} We start by comparing the feature importances natively provided by FLANs against three gradient-based \emph{post-hoc} feature attribution methods: Integrated Gradients \citep{Sundararajan2017AxiomaticNetworks}, Saliency \citep{Simonyan2013DeepMaps}, and InputXGradient \citep{Kindermans2016InvestigatingNetworks}. Figure~\ref{fig:cub_attr_comparison} shows the importances computed by the aforementioned methods for a test sample correctly predicted as a \texttt{Black Footed Albatross}. The norms of the latent representations (Section~\ref{sec:feature_imp}) highlight features that are typically used to identify birds, \ie the region around the eye and the beak \citep{Chen2019ThisRecognition}. While noisier, gradient-based methods highlight some of the same regions, partially validating our model. However, IntegratedGradients and Saliency also highlight some areas in the top-right part of the image. In the next paragraph, we argue that these regions might be indicative of \emph{counterfactual} evidence, rather than ``direct'' evidence.

\begin{figure}[h]
    \centering
    \includegraphics[width=\linewidth]{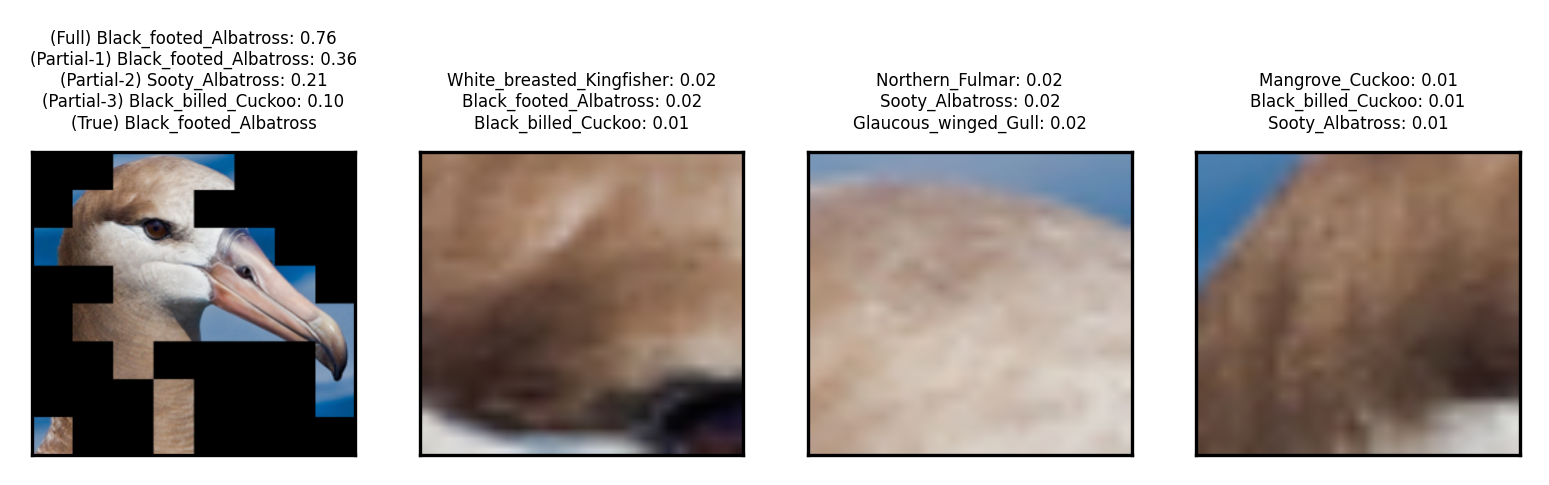}
    
    \caption{Algorithmic Interpretation of FLANs. In the leftmost image, we show the top 20 most important patches for a \texttt{Black Footed Albatross}. On top of the image we report the top 3 predicted classes (with probabilities) computed on the shown partial image. For reference, we provide also the prediction on the full image and the true label. In the 3 rightmost images we show the top 3 most important patches, together with the top 3 predictions for each of them.}
    \label{fig:cub_interpret_patches}

\end{figure}

\paragraph{Algorithmic Interpretation.} To show that the top-right area of the image is not particularly important towards the prediction of the sample at hand, we leverage the algorithmic interpretation modality described in Section~\ref{sec:interpret_mech}. More precisely, we drop from the inner sum of Eq.~\eqref{eq:lanet} those features with the smallest latent norm. This is shown in Figure~\ref{fig:cub_interpret_patches} (left). Note that this is different from occluding the image \citep{Zeiler2014VisualizingNetworks}.
As shown at the top of Figure~\ref{fig:cub_interpret_patches} (left), the incomplete image is still classified as \texttt{Black Footed Albatross}. This makes sense since, as argued in Section~\ref{sec:feature_imp}, features with small norm do not contribute to the sample representation and, consequently, to the prediction. This confirms that the top-right area is not important as direct evidence towards \texttt{Black Footed Albatross}. Rather it may be an area that, if modified, it may \emph{become} important towards a different classification, \ie a counterfactual. 

To further demonstrate the usefulness of separability towards interpretability, we note that the incomplete image has a 10\% chance to be classified as \texttt{Black Billed Cuckoo}. To see why it may be, we can apply the prediction network on the single patches. The 3 rightmost images in Figure~\ref{fig:cub_interpret_patches} show how the top 3 patches are classified. In particular, we note that the first and third patch provide evidence towards \texttt{Black Billed Cuckoo}.

\begin{figure}[h]
    \centering
    \includegraphics[width=\linewidth]{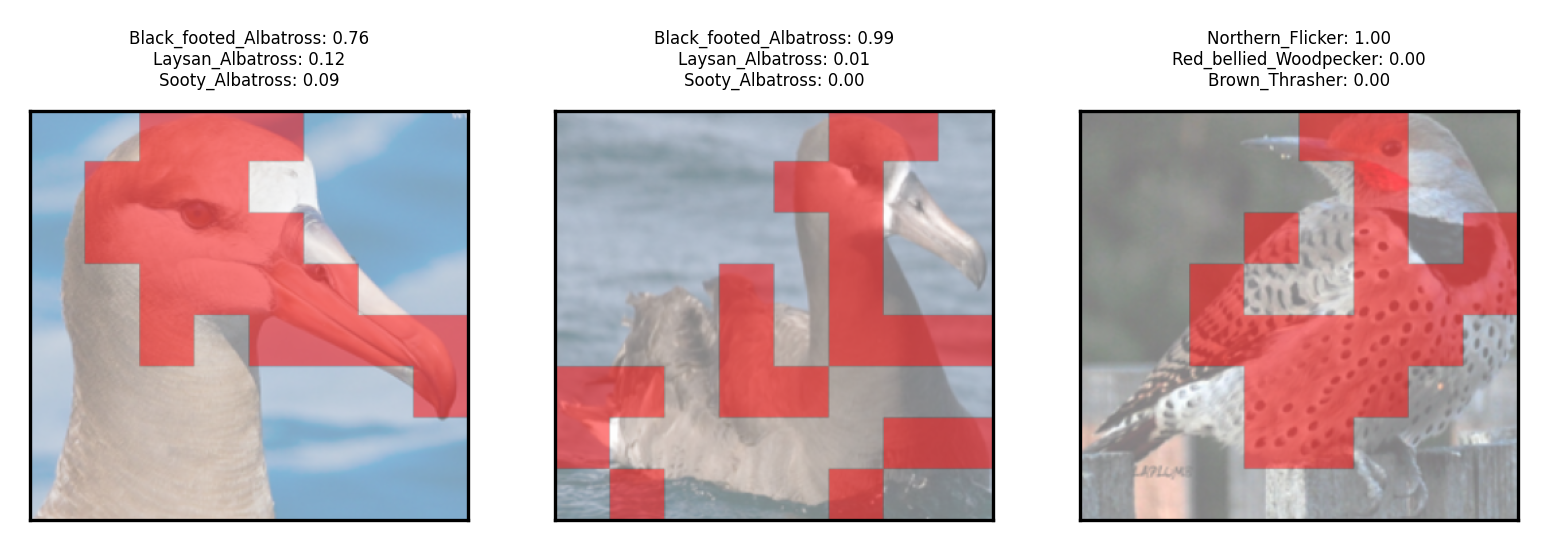}
    
    \caption{Example-Based Interpretation of FLANs. (Left) Reference image. (Center) Closest image in the latent space. (Right) Farthest image in the latent space. Above each image, we report the top 3 predicted classes (with probabilities). We highlight in red the patches that explain \emph{why} the center (resp. right) image is close (resp. far) from the reference image.}
    \label{fig:cub_interpret_example}
\end{figure}

\paragraph{Example-Based Explanations.} We finally show the third modality for interpreting FLANs, \ie by examples. As discussed in Section~\ref{sec:example_based}, to do this we need to look for the nearest neighbors in the latent space $\mathcal{Z}$ before the prediction network. While it can be argued that any layer in the prediction network can be used as a sample representation for nearest neighbor search, we argue that the representation before the prediction network is the most representative of the sample. This is because this is the first point where the information from all the features is aggregated, and no further processing has been performed yet (cf. data processing inequality, \eg \citealt{Cover2006ElementsProcessing}). Figure~\ref{fig:cub_interpret_example} shows the bird (center) that most closely resembles the reference \texttt{Black Footed Albatross} (left). In particular, for these images, the top 3 classes predicted are the same (reported on top of the images). However, it can be noted that the computed probabilities for each of these classes are different for the two birds. This means that the two images contain somewhat different information. It is then reasonable to wonder what information do the two images have in common, \ie why are these two images similar/different? This can be an important question for assessing similarity models \citep{Eberle2020BuildingModels}. With FLANs, this question can be answered by means of a \emph{linear assignment problem}~\citep{Burkard1999LinearExtensions}. More details are provided in Appendix~\ref{app:lin_assign}.

The red patches in the central image of Figure~\ref{fig:cub_interpret_example} are the \emph{most similar} patches assigned to the top features of the reference image on the left. These results show that, again, the eyes of a bird are a distinctive feature. In particular, this feature seems to drive the similarity between the two images. Analogously, if we look for the \emph{most dissimilar} patches between the reference image and the most dissimilar training sample (rightmost image in Figure~\ref{fig:cub_interpret_example}), we find that the eyes (together with the feathers) are used to distinguish the \texttt{Black Footed Albatross} from the \texttt{Northern Flicker}.

\subsection{Quantitative interpretability evaluation}

Quantitatively assessing a subjective matter, such as interpretability, can be cumbersome. One of the ways for this quantitative evaluation is  what \citet{Doshi-Velez2017TowardsLearning} call \emph{functionally-grounded} evaluation. In particular, we leverage the recent work by \citet{Nguyen2020OnInterpretability}, where the authors propose a series of metrics to assess the \emph{faithfulness} of an explanation (\ie how ``correct'' is the explanation) and its broadness (\ie how general it is).

For feature attributions, the faithfulness is partially assessed by \emph{monotonicity} (computed as how unpredictable the sample becomes if important features are removed) and by \emph{non-sensitivity} (\ie unimportant features should not be given a high score). For example-based methods, faithfulness is given by the (inverse) \emph{non-representativeness}, where representativeness means that the similar samples should be classified similarly. Broadness, instead, is given by the \emph{diversity} of the examples. 

In Table~\ref{tab:fm}, we compare the attribution scores natively provided by FLAN as the norms of the feature latent representations against the same methods mentioned in Section~\ref{sec:exp_cub}. In Table~\ref{tab:ex_metrics}, we use the example-based metrics to evaluate if the feature latent space (LS) is better than the original input space (OS) for the selection of (12) prototypes via K-Medoids~\citep{Kaufman2009FindingAnalysis}.The same analysis can be done using a different method such as ProtoDash~\citep{Gurumoorthy2017EfficientWeights}. Note that we evaluate example-based strategies in both a local and global setting. In the global, we select examples to explain whole classes, while in the local we want to explain individual test samples using their nearest training samples.


\begin{table*}[h]
\caption{Comparison of FLAN norms, Integrated Gradients (IG), InputXGradient (InpXGrad), and Saliency using feature attribution metrics.}\label{tab:fm}

    \begin{subtable}[h]{0.48\linewidth}
    \centering
    \caption{Monotonicity. Higher is better.}\label{monot}
    
    \resizebox{\linewidth}{!}{
    \begin{tabular}{lccccc}
    \toprule
         & \texttt{TCR-Epitope}  & \texttt{MNIST} & \texttt{CUB} & \texttt{SVHN} & \texttt{AGNews} \\
    \midrule
    FLAN & \textbf{0.17} & 0.10 & 0.02 & \textbf{0.04} & \textbf{0.17} \\ 
    IG & 0.07 & 0.15 & 0.03 & \textbf{0.04} & 0.16 \\
    InpXGrad & 0.07  & \textbf{0.16} & 0.04 & 0.03 & 0.15 \\
    Saliency & 0.10 & 0.15 & \textbf{0.06} & \textbf{0.04} & 0.09 \\
    \bottomrule
    \end{tabular}
    }
    
    \end{subtable}
    \hfill
    \begin{subtable}[h]{0.48\linewidth}
    \centering
    \caption{Non-Sensitivity. Lower is better.}\label{nsens}
    
    \resizebox{\linewidth}{!}{
    \begin{tabular}{lccccc}
    \toprule
         & \texttt{TCR-Epitope}  & \texttt{MNIST} & \texttt{CUB} & \texttt{SVHN} & \texttt{AGNews}\\
    \midrule
    FLAN & 52.89 & 483 & 1886.00 & 27.29 & 11.70\\ 
    IG & 25.19 & \textbf{1.83} & \textbf{5.40} & 4.57 & 12.58\\
    InpXGrad & 24.04 & 3.50 & 5.80 & \textbf{5.14} & \textbf{10.59}\\
    Saliency & \textbf{23.50} & 168.85 & 111501.20 & 1015.44 & 42.19\\
    \bottomrule
  \end{tabular}
  }
  
  \end{subtable}
  
\end{table*}

\begin{table*}[h]
    \centering
    \caption{Comparison of examples/prototypes computed by clustering in the Original input Space (OS) and the feature Latent Space (LS) in both the local and global setting. Comparison by diversity (D, higher is better) and non-representativeness (NR, lower is better).}\label{tab:ex_metrics}\label{ebm}
    \resizebox{0.9\linewidth}{!}{
    \begin{tabular}{lcccccccccc}
        \toprule
        & \multicolumn{2}{c}{\texttt{TCR-Epitope}} & \multicolumn{2}{c}{\texttt{MNIST}} & \multicolumn{2}{c}{\texttt{CUB}}  & \multicolumn{2}{c}{\texttt{SVHN}} & \multicolumn{2}{c}{\texttt{AGNews}}\\
    \cmidrule(l){2-3}\cmidrule(l){4-5}\cmidrule(l){6-7}\cmidrule(l){8-9}\cmidrule(l){10-11}
    & OS & LS & OS & LS & OS & LS & OS & LS & OS & LS\\
         
    \midrule
    Local D & \textbf{299.27} & 51.85 & \textbf{68.35} & 66.09 & \textbf{1507.18}  & 1187.72 & 67.23 & \textbf{203.82} & 0.51 & \textbf{0.53}\\
    Local NR & \textbf{2.21} & 2.39 & \textbf{2.15}  & 2.19 & 3.63  & \textbf{3.47} & \textbf{2.14} & 2.13 & 0.87 & \textbf{0.10} \\
    \midrule
    Global D & \textbf{671.46} & 67.74 & \textbf{61.01} & 60.49  & \textbf{1376.06} & 1332.07 & 185.15 & \textbf{190.12} & 0.31 & \textbf{0.48} \\
    Global NR & \textbf{0.18} & 0.25 & \textbf{2.16} & 2.17 & \textbf{3.81} & 3.83 & \textbf{2.09} & \textbf{2.09} & 7e-3 & \textbf{6e-4}\\
    \bottomrule
  \end{tabular}
  }
  \end{table*}

The non-sensitivity results show that FLAN tends to either assign some importance to not influential features, or no importance to influential features. This can be explained if we note that the reported non-sensitivity is particularly unfavorable for patch-based FLANs in image classification: to make the attribution methods comparable we needed an attribution scores for \emph{each pixel}, and therefore for FLAN scores we assigned the same value to each pixel in a given patch, even though certain pixels may not actually have any influence. This also explains the relatively low, but comparable monotonicity performance of FLAN scores on the image datasets. However, in non-patch-based datasets, FLAN scores seem to perform best. Considering the above points, overall FLAN performs satisfactorily, showing that to FLANs interpret via attribution analysis there is \emph{no need to rely on post-hoc methods}.

The example-based metrics (Table~\ref{tab:ex_metrics}) show that there does not seem to be a clear advantage between clustering in the original input space or in the latent space, apart for more extreme cases (\eg \texttt{TCR-Epitope} and \texttt{SVHN}). This suggests that an user may simply apply example-based strategies directly to the input space. However, we argue that example-based analysis in the feature space may turn out useful in two instances. Firstly, FLAN feature spaces offer the possibility to identify which features drive the (dis-)similarity of the selected prototypes (Section~\ref{sec:exp_cub}). Secondly, sometimes using the euclidean distance in the feature space may reveal easier than either defining or computing a distance in the original input space (\eg when dealing with graphs), therefore offering a computational advantage.

\section{CONCLUSION}

In this paper, we introduced a class of \emph{powerful and interpretable} models, which we call FLANs. In terms of prediction accuracy, our results show that our model can achieve reasonable performance on \emph{complex datasets}. We demonstrate how our model can be easily and natively interpreted \emph{without the need of post-hoc methods} and \emph{without major drawbacks in terms of interpretability} (according to recently introduced metrics).

We believe that our work represents an important contribution towards interpretable machine learning in \emph{complex domains} with profound societal impact, \eg histopathology and healthcare. In future work, we plan to address the generalization difficulties of our model. Moreover, we plan to run user studies to further validate the interpretability of our model.

\subsubsection*{Acknowledgements}

We acknowledge funding by the European Union’s Horizon
2020 research and innovation programme (iPC–Pediatric
Cure, No. 826121).

\bibliographystyle{unsrtnat}
\bibliography{references}

\onecolumn
\aistatstitle{Supplementary Materials}

\section{Training details}\label{app:train_det}

All the deep learning models were trained using \texttt{Adam} \citep{Kingma2015Adam:Optimization} (or variants thereof, \ie \texttt{AdamW} \citealt{Loshchilov2017DecoupledRegularization}, \texttt{RAdam} \citealt{Liu2020OnBeyond}). Learning rates varied in the set $\{0.001, 0.0005, 0.0001, 0.00005\}$. Training was tested with no learning rate scheduling, as well as exponential decay, step decay, and cosine annealing \citep{Loshchilov2016SGDR:Restarts} (with and without restart). The chosen hyperparameters for each experiment can be retrieved from the corresponding \texttt{config.json} file provided in the accompanying code. Details about the architectures used are also provided in the accompanying code.

\paragraph{Tabular datasets.} For each datasets, 10 different runs are performed. Table~\ref{tab:tab_benchmark} reports means among these runs. In Table~\ref{tab:tab_benchmark_app} we report again the mean, together with standard deviation and top performance (max).

\paragraph{Image datasets.} For each datasets, 5 different runs are performed (with the best set of hyperparameters). Table~\ref{tab:img_benchmark} reports top performance among these runs. In Table~\ref{tab:img_benchmark_app} we report again the top performance (max), together with the mean among the 5 runs, and standard deviation.

\paragraph{Text datasets.} For each datasets, 5 different runs are performed (with the best set of hyperparameters). Table~\ref{tab:text_benchmark} reports top performance among these runs. In Table~\ref{tab:text_benchmark_app} we report again the top performance (max), together with the mean among the 5 runs, and standard deviation.

\paragraph{TCR-epitope dataset.} For this dataset 100 runs are performed (with the best set of hyperparameters). In a second setting, we add a pre-training step that learns the embedding representation separately for the TCR and epitope sequences. In this step, we utilize publicly available datasets containing only TCR or epitope sequences to find good representations and further improve FLAN's ability to focus on meaningful positions/amino-acids. After pre-training, FLAN is implemented as usual. For this setting we once more performed 100 runs (with the best set of hyperparameters).

\section{Detecting similar and different information content}\label{app:lin_assign}

While analyzing an example-based interpretation, we may wonder why two samples are similar/different. This can be answered relatively easily by leveraging the FLAN architecture. Informally, we can reformulate this question in the following way. Given the top $K$ features of a sample $\mathbf{x}$, we would like to find the $L$ features in the second sample $\mathbf{\hat{x}}$ that most closely match them. Mathematically,
\begin{equation}
   \min_{S \subset \{1 \dots N\}, |S| = L} || \sum_{i \in T} \mathbf{z}_i - \sum_{j \in S}\mathbf{\hat{z}}_j||
\end{equation}

where $T$ (with $|T|=K$) is the set of the top $K$ features $\mathbf{z}_i$ of sample $\mathbf{x}$, and $S$ is any subset of $\{1 \dots N\}$ of cardinality $L$. To simplify the problem, we restrict $L$ to be equal to $K$. We further leverage the triangle inequality to split the norm of the sum in a sum of norms. We therefore end up with the following \emph{linear assignment problem} \citep{Burkard1999LinearExtensions} with assignment function $A: T \rightarrow \{1,\dots,N\}$:
\begin{equation}
   \min_A \sum_{i \in T} || \mathbf{z}_i - \mathbf{\hat{z}}_{A(i)}||
\end{equation}

\section{Further results}\label{app:more_res}

\subsection{Performance results}\label{app:more_benchmark}

Here we report more detailed performance results (Table~\ref{tab:benchmark_app}). The experimental setup is reported in the previous section~\ref{app:train_det}. Note that the inconsistencies in the results between Table~\ref{tab:benchmark} and Table~\ref{tab:benchmark_app} are due to the fact that new experiments have been run with \emph{not} fixed random seeds to produce Table~\ref{tab:benchmark_app}. However, this change does not affect the interpretation of the results discussed in Section~\ref{sec:perf_results}.

\begin{table}[h!]
  \caption{Further benchmarking results. (*) denotes pretrained models or models using automated augmentation strategies. (**) denotes FLAN models that use pretrained models as \emph{part} of the feature functions.  Reporting mean (max) $\pm$ standard deviation over 10 runs for tabular datasets, and 5 runs for the other datasets.}\label{tab:benchmark_app}
  \begin{subtable}{\textwidth}
  \centering
  \caption{Area under the curve (AUC) on tabular datasets.}\label{tab:tab_benchmark_app}
  \begin{tabular}{lcccc}
    \toprule
         & \texttt{COMPAS}  & \texttt{adult} & \texttt{heart} & \texttt{mammo} \\
    \midrule
    \multirow{2}{*}{Logistic Regression 
    } & 0.905 (0.917) & 0.892 (0.896) & 0.873 (0.923) & \textbf{0.841} (0.874) \\
    & $\pm$ 0.006 & $\pm$ 0.003 & $\pm$ 0.032 & $\pm$ 0.017\\
    \multirow{2}{*}{Decision Tree (small) 
    }& 0.903 (0.915) & 0.865 (0.871) & 0.849 (0.882) & 0.799 (0.818) \\
    & $\pm$ 0.007 & $\pm$ 0.005 & $\pm$ 0.026 & $\pm$ 0.017\\
    \multirow{2}{*}{Decision Tree (unrestricted) 
    } & 0.902 (0.915) & 0.813 (0.821) & 0.848 (0.882) & 0.801 (0.826) \\
    & $\pm$ 0.007 & $\pm$ 0.005 & $\pm$ 0.024 & $\pm$ 0.016\\
    \multirow{2}{*}{Random Forest 
    }  & \textbf{0.915} (0.927) & 0.869 (0.877) & 0.945 (0.964) & 0.822 (0.841) \\
    & $\pm$ 0.007 & $\pm$ 0.004 & $\pm$ 0.014 & $\pm$ 0.016\\
    \multirow{2}{*}{EBM 
    }  & 0.911 (0.923) & \textbf{0.893} (0.896) & 0.941 (0.959) & 0.840 (0.869)  \\
    & $\pm$ 0.008 & $\pm$ 0.002 & $\pm$ 0.015 & $\pm$ 0.015\\
    \multirow{2}{*}{MLP}  & \textbf{0.915} (0.927) & 0.874 (0.883) & 0.937 (0.958) &  0.831 (0.856) \\
    & $\pm$ 0.006 & $\pm$ 0.005 & $\pm$ 0.023 & $\pm$ 0.014\\
    \multirow{2}{*}{SENN \citep{AlvarezMelis2018TowardsNetworks}}  & 0.910 (0.922) & 0.865 (0.873) & 0.881 (0.925) &  0.834 (0.860) \\
    & $\pm$ 0.007 & $\pm$ 0.005 & $\pm$ 0.036 & $\pm$ 0.013\\
    \midrule
    \multirow{2}{*}{FLAN}  & 0.914 (0.923) & 0.880 (0.886) & \textbf{0.950} (0.973) &  0.832 (0.867) \\
    & $\pm$ 0.004 & $\pm$ 0.004 & $\pm$ 0.019 & $\pm$ 0.019\\
    \bottomrule
  \end{tabular}
  \end{subtable}
  
  \vspace{1ex}
  
  \begin{subtable}{.55\linewidth}
  \centering
  \caption{Test accuracy (\%) on image datasets.}\label{tab:img_benchmark_app}
  \begin{tabular}{lccc}
    \toprule
         & \texttt{MNIST} & \texttt{SVHN} & \texttt{CUB} \\
    \midrule
    ResNet 
    & 99.2 & 94.5* & 84.5* \\ 
    iCaps 
    & 99.2 & 92.0 & - \\
    ViT 
    & - & 88.9 & 90.4* \\
    ProtoPNet 
    & - & - & 84.8*\\
    SENN 
    &  99.1 &  - & - \\
    SotA 
    & \textbf{99.84} & \textbf{99.0}* & \textbf{91.3}*\\
    \midrule
    \multirow{3}{*}{FLAN}  & 99.00 & 93.37 & 71.17\\
    & (99.05) & (93.41) & (71.53)\\
    & $\pm$ 0.0007 & $\pm$ 0.0004 & $\pm$ 0.003\\
    \bottomrule
  \end{tabular}
  \end{subtable}
  \hfill
  \begin{subtable}{.4\linewidth}
  \centering
  \caption{Test accuracy (\%) on text datasets.}\label{tab:text_benchmark_app}
  \begin{tabular}{lcc}
    \toprule
         & \texttt{AGNews} & \texttt{IMDb} \\
    \midrule
    CharCNN 
    & 90.49 & - \\
    LSTM 
    & 93.8 & 86.5 \\
    VDCNN 
    & 91.33 & 79.47 \\
    HAHNN 
    & - & 95.17 \\
    XLNet 
    & \textbf{95.6}* & \textbf{96.8}* \\
    \midrule
    \multirow{3}{*}{FLAN}  & 90.6 & 84.9  \\
      & (90.9) & (85.1)  \\
    & $\pm$ 0.003 & $\pm$ 0.002 \\
    \bottomrule
  \end{tabular}
  \end{subtable}
\end{table}

Table \ref{pf} summarizes the performance results for the two settings in the TCR-epitope dataset. Even though the performance metrics are almost identical, we argue that pretrained FLAN concentrates on more meaningful positions, since the amino-acid embedding representations are learned with more data.

\begin{table}[h!]
    \centering
    \caption{ROC-AUC and Balanced accuracy for FLAN and pretrained FLAN in the TCR-epitope dataset.}
    \label{pf}
    \begin{tabular}{lcc}
    \toprule
    & AUC & B.Accuracy \\
    \midrule
     FLAN & 0,862 $\pm$0,006  & 0,783 $\pm$0,007 \\
     pretrained FLAN & 0,858 $\pm$0,004 & 0,780 $\pm$0,006\\
     \bottomrule
    \end{tabular}

\end{table}

\clearpage

\subsubsection{Training graphs}

\begin{figure}[h!]
     \centering
     \begin{subfigure}[b]{0.45\textwidth}
         \centering
         \includegraphics[width=\textwidth]{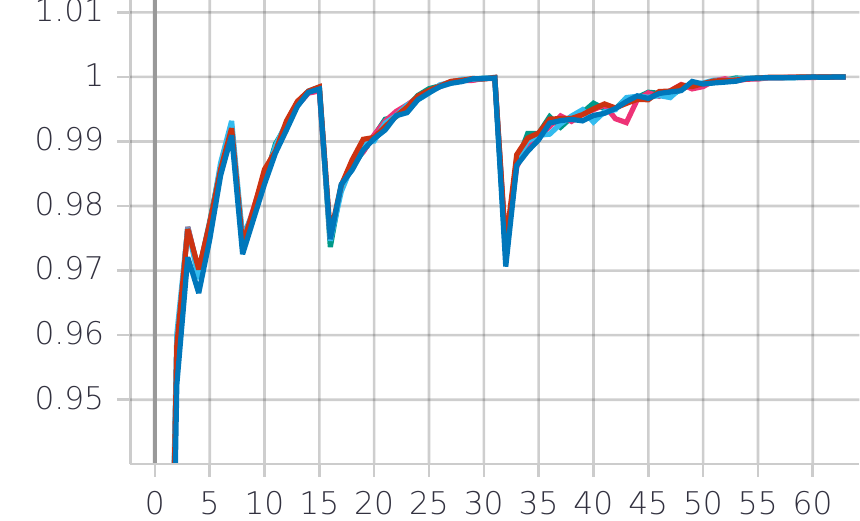}
         \caption{Training graphs for the \texttt{MNIST} dataset for 5 experiments (top \emph{train} performance $100\%$).}
         \label{fig:mnist_train}
     \end{subfigure}
     \hfill
     \begin{subfigure}[b]{0.45\textwidth}
         \centering
         \includegraphics[width=\textwidth]{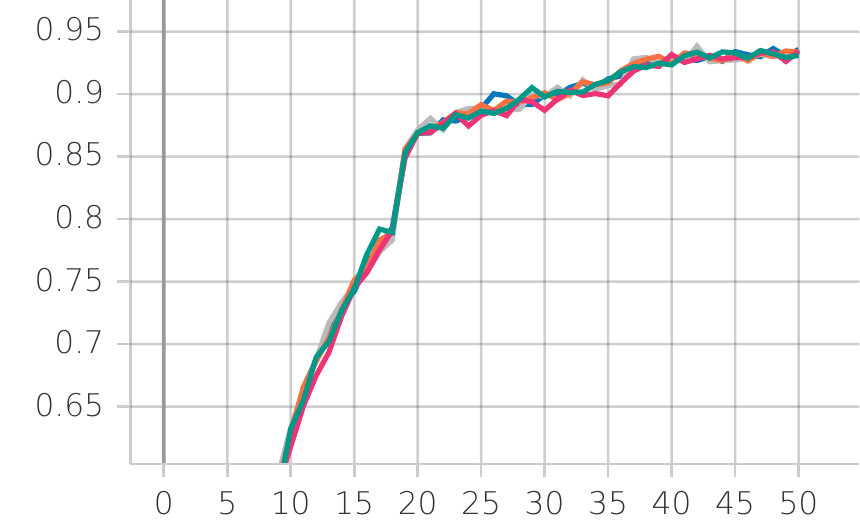}
         \caption{Training graphs for the \texttt{CUB} dataset for 5 experiments (top \emph{train} performance $94\%$).}
         \label{fig:cub_train}
     \end{subfigure}
     
     \begin{subfigure}[b]{0.45\textwidth}
         \centering
         \includegraphics[width=\textwidth]{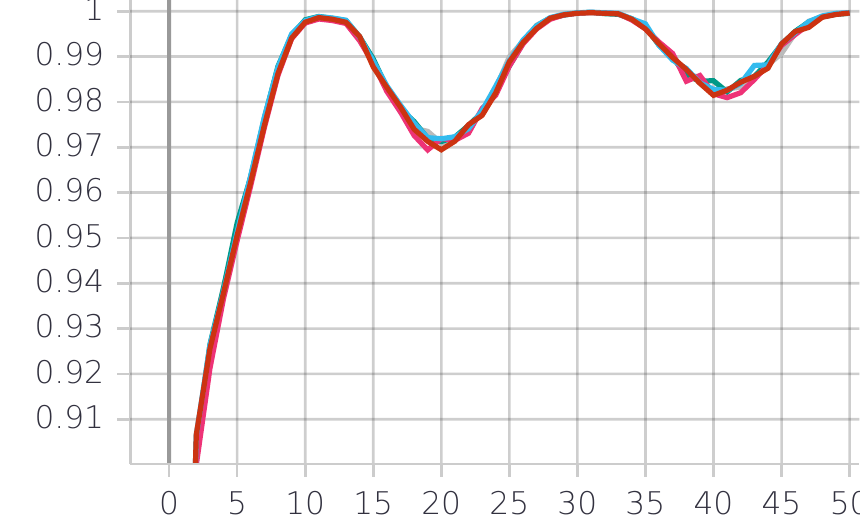}
         \caption{Training graphs for the \texttt{AGNews} dataset for 5 experiments (top \emph{train} performance $100\%$).}
         \label{fig:agnews_train}
     \end{subfigure}
     \hfill
     \begin{subfigure}[b]{0.45\textwidth}
         \centering
         \includegraphics[width=\textwidth]{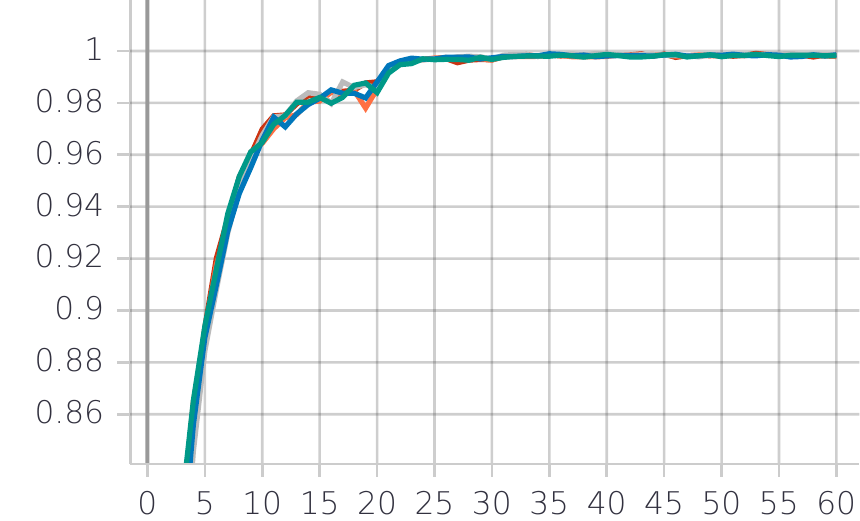}
         \caption{Training graphs for the \texttt{IMDb} dataset for 5 experiments (top \emph{train} performance $100\%$).}
         \label{fig:imdb_train}
     \end{subfigure}
        \caption{Training graphs for various experiments.}
        \label{fig:train_graphs}
\end{figure}

In Section~\ref{sec:perf_results} and Section~\ref{app:more_benchmark} we reported the best models in terms of \emph{test} performance. Here we report (Figure~\ref{fig:train_graphs}) the training curves of some models that achieved $100\%$ (resp. $94\%$) training accuracy on \texttt{MNIST}, \texttt{AGNews}, and \texttt{IMDB} (resp. \texttt{CUB}), but generalised poorly.

\subsection{Interpretability results}\label{app:more_interpretability}

\subsubsection{\texttt{CUB}}

Here we show more interpretability results from the \texttt{CUB} dataset.

\begin{figure}[h!]
     \centering
     \begin{subfigure}[b]{\textwidth}
         \centering
         \includegraphics[width=\textwidth]{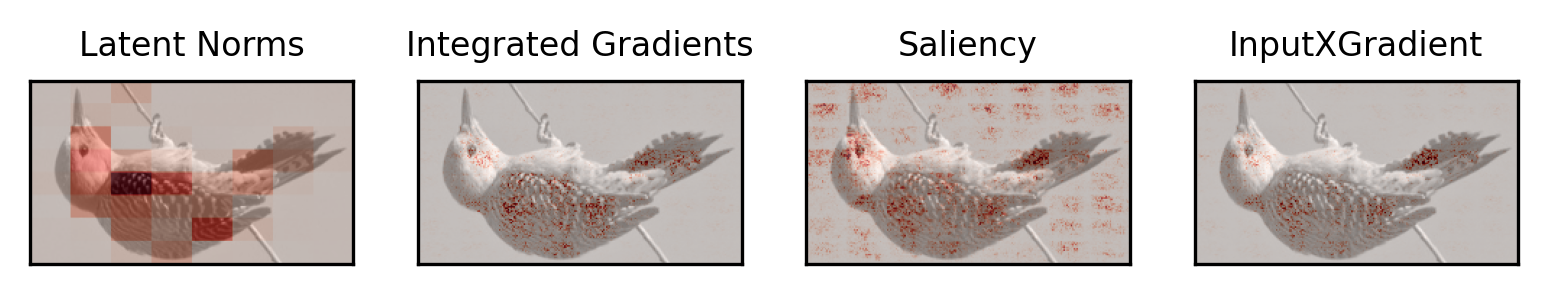}
         \caption{Feature attribution comparisons.}
         \label{fig:feat_attr_more_comparision_1}
     \end{subfigure}
     
     \begin{subfigure}[b]{\textwidth}
         \centering
         \includegraphics[width=\textwidth]{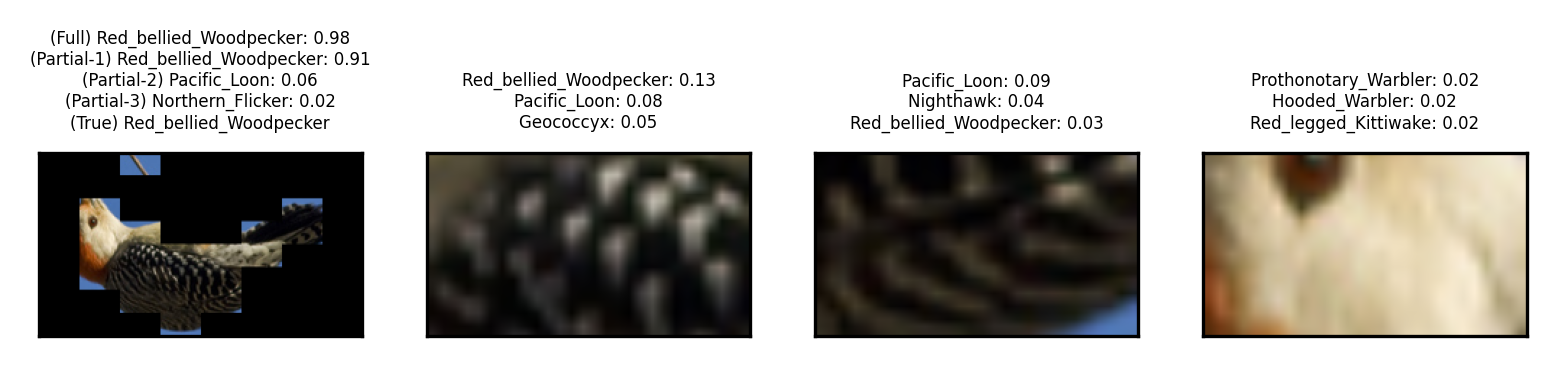}
         \caption{Algorithmic interpretation.}
         \label{fig:algo_more_comparision_1}
     \end{subfigure}

     \begin{subfigure}[b]{\textwidth}
         \centering
         \includegraphics[width=\textwidth]{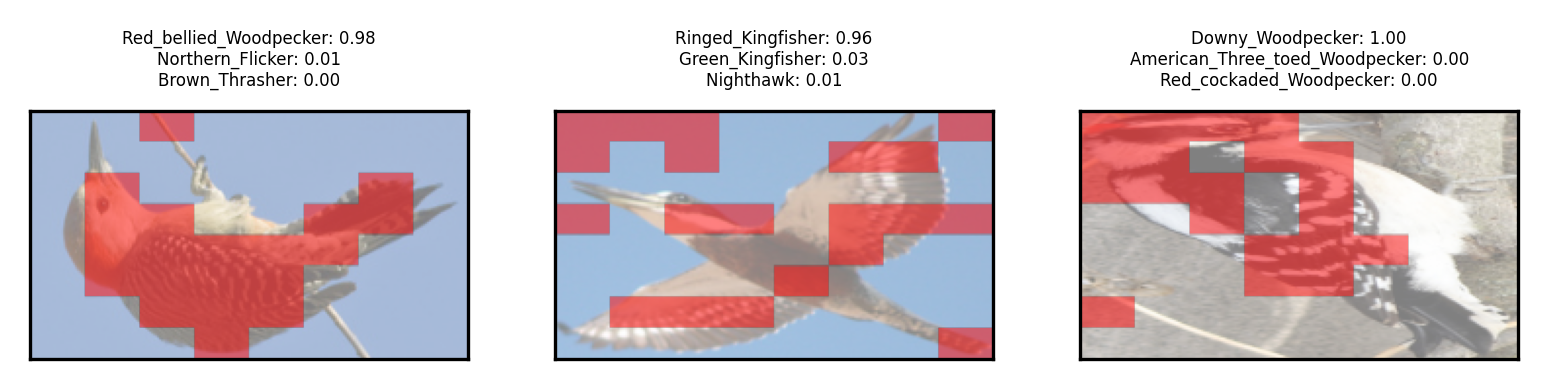}
         \caption{Example-based interpretation with correspondences.}
         \label{fig:example_more_comparision_1}
     \end{subfigure}
        \caption{Additional interpretability results.}
        \label{fig:app_cub_1}
\end{figure}

\begin{figure}[h!]
     \centering
     \begin{subfigure}[b]{\textwidth}
         \centering
         \includegraphics[width=\textwidth]{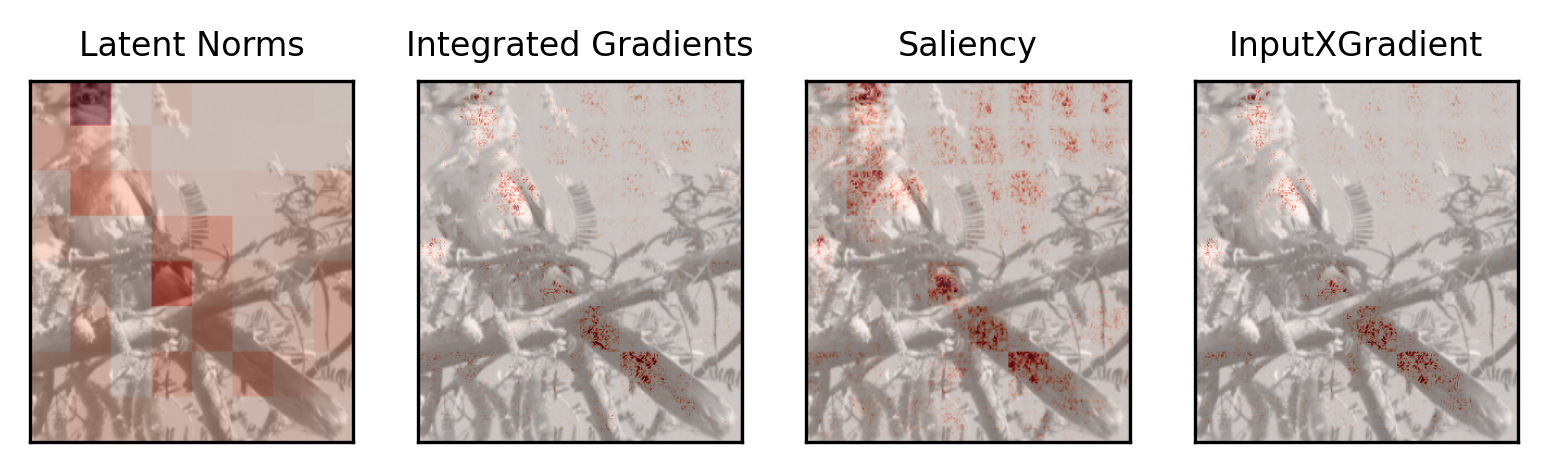}
         \caption{Feature attribution comparisons.}
         \label{fig:feat_attr_more_comparision_2}
     \end{subfigure}
     
     \begin{subfigure}[b]{\textwidth}
         \centering
         \includegraphics[width=\textwidth]{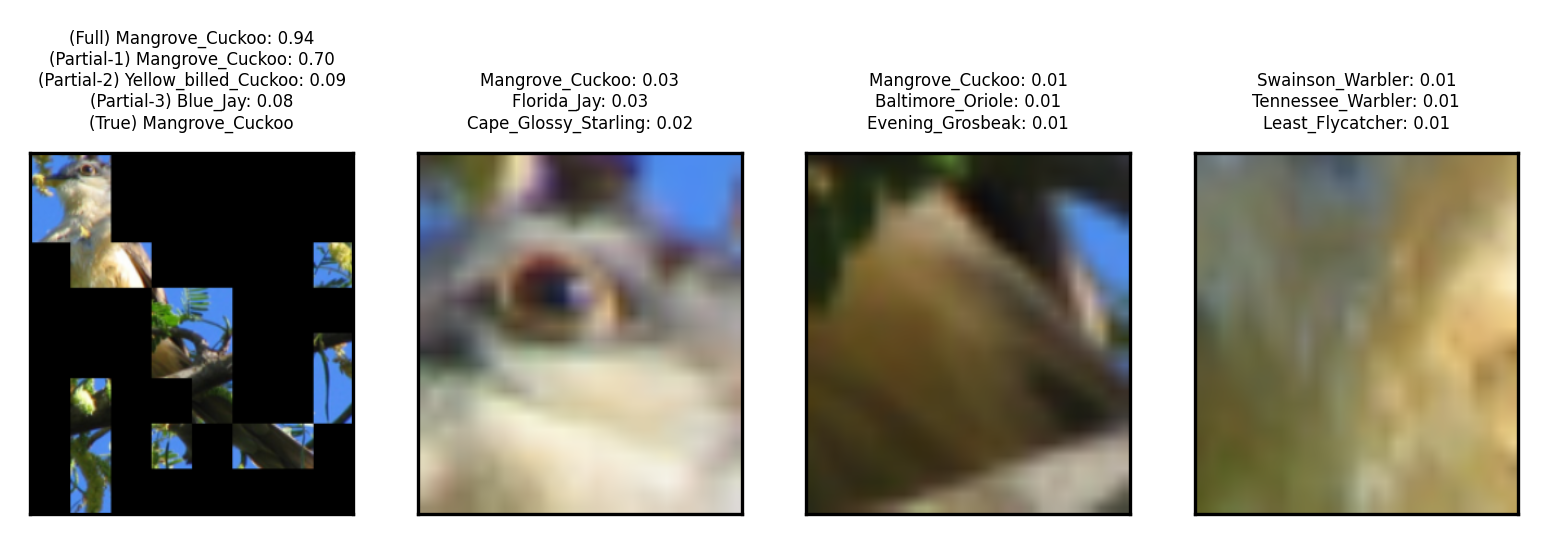}
         \caption{Algorithmic interpretation.}
         \label{fig:algo_more_comparision_2}
     \end{subfigure}

     \begin{subfigure}[b]{\textwidth}
         \centering
         \includegraphics[width=\textwidth]{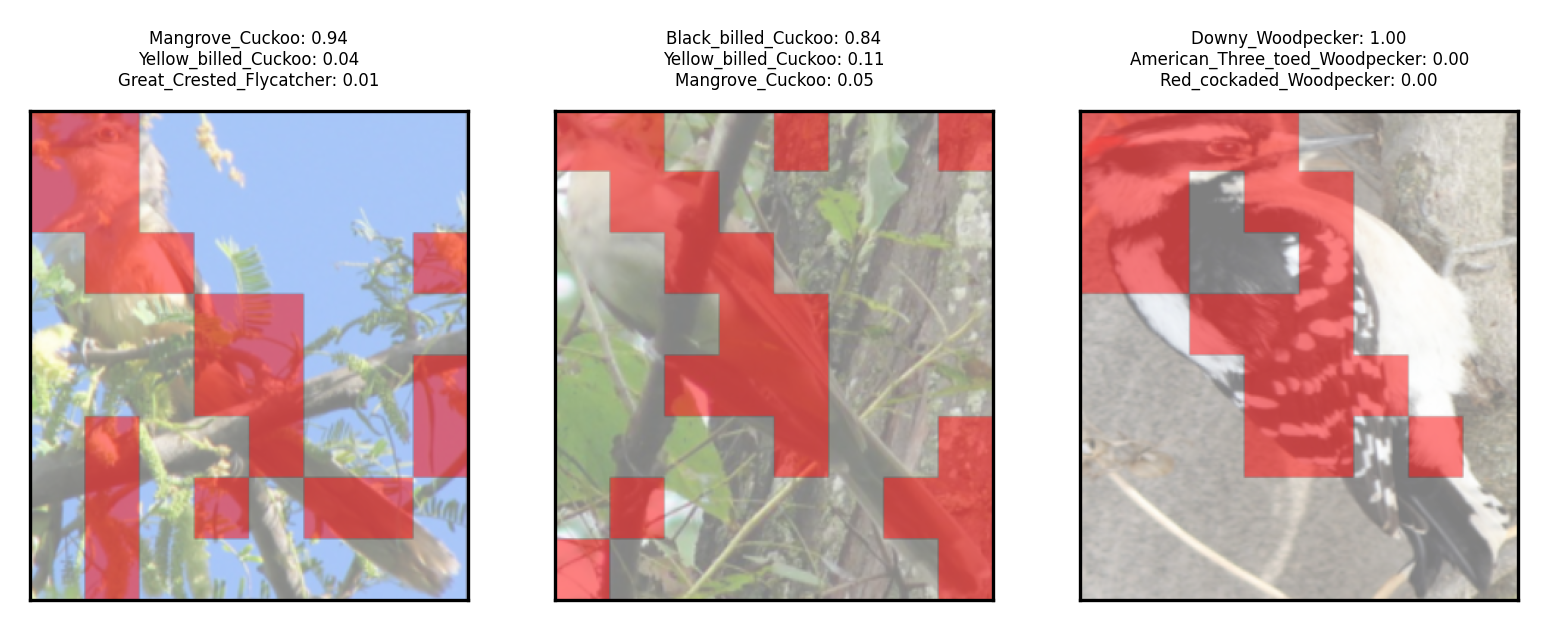}
         \caption{Example-based interpretation with correspondences.}
         \label{fig:example_more_comparision_2}
     \end{subfigure}
        \caption{Additional interpretability results.}
        \label{fig:app_cub_2}
\end{figure}

\begin{figure}[h!]
     \centering
     \begin{subfigure}[b]{\textwidth}
         \centering
         \includegraphics[width=\textwidth]{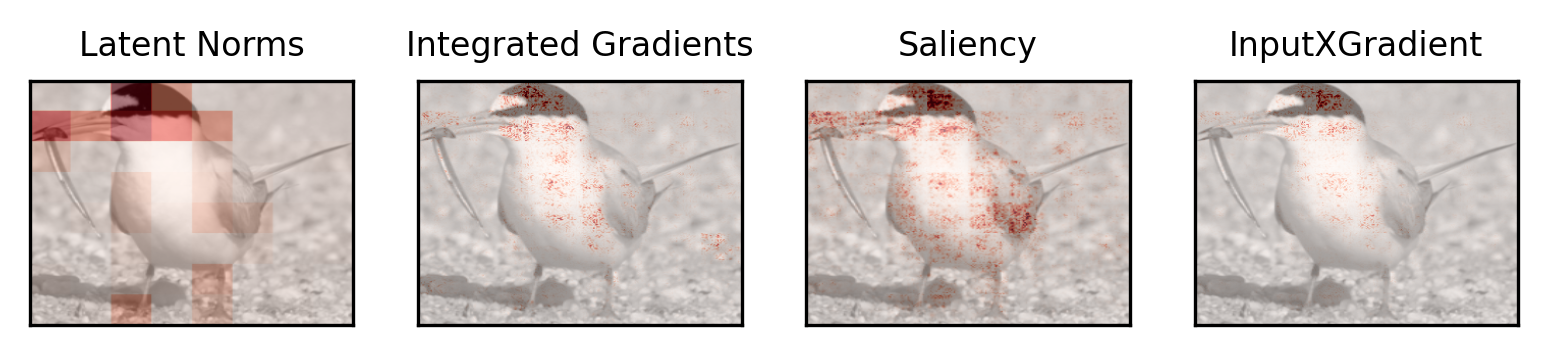}
         \caption{Feature attribution comparisons.}
         \label{fig:feat_attr_more_comparision_3}
     \end{subfigure}
     
     \begin{subfigure}[b]{\textwidth}
         \centering
         \includegraphics[width=\textwidth]{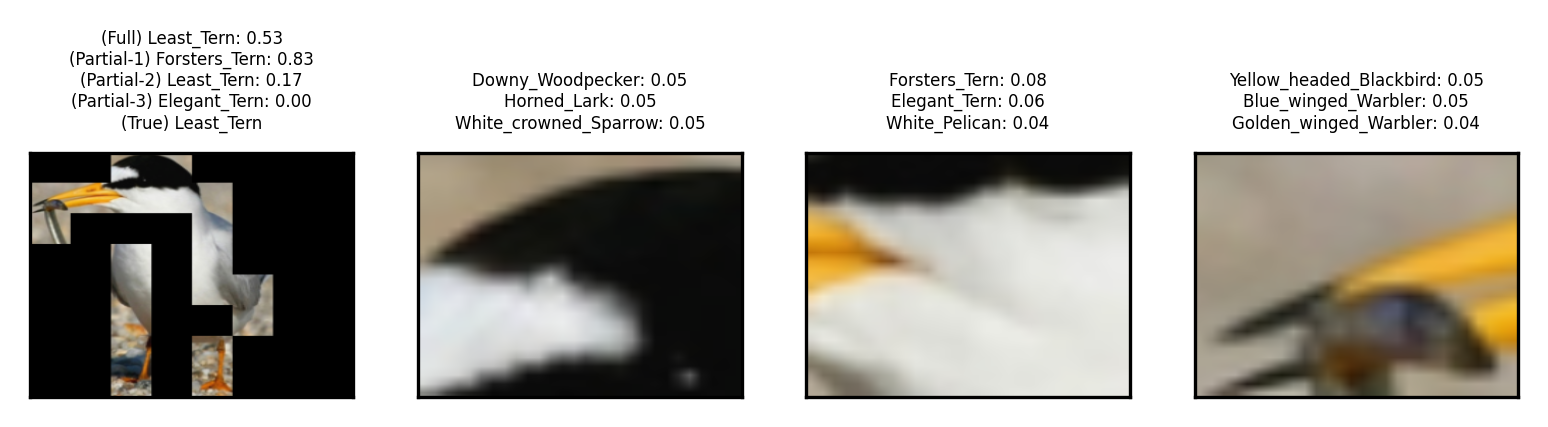}
         \caption{Algorithmic interpretation.}
         \label{fig:algo_more_comparision_3}
     \end{subfigure}

     \begin{subfigure}[b]{\textwidth}
         \centering
         \includegraphics[width=\textwidth]{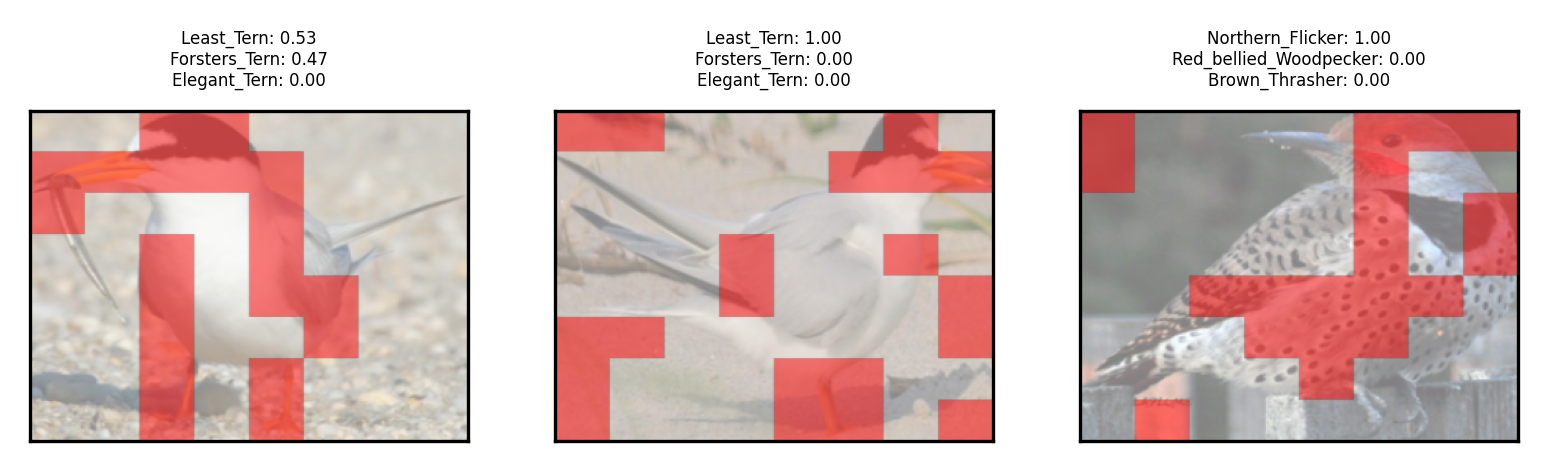}
         \caption{Example-based interpretation with correspondences.}
         \label{fig:example_more_comparision_3}
     \end{subfigure}
        \caption{Additional interpretability results.}
        \label{fig:app_cub_3}
\end{figure}

\begin{figure}[h!]
     \centering
     \begin{subfigure}[b]{\textwidth}
         \centering
         \includegraphics[width=\textwidth]{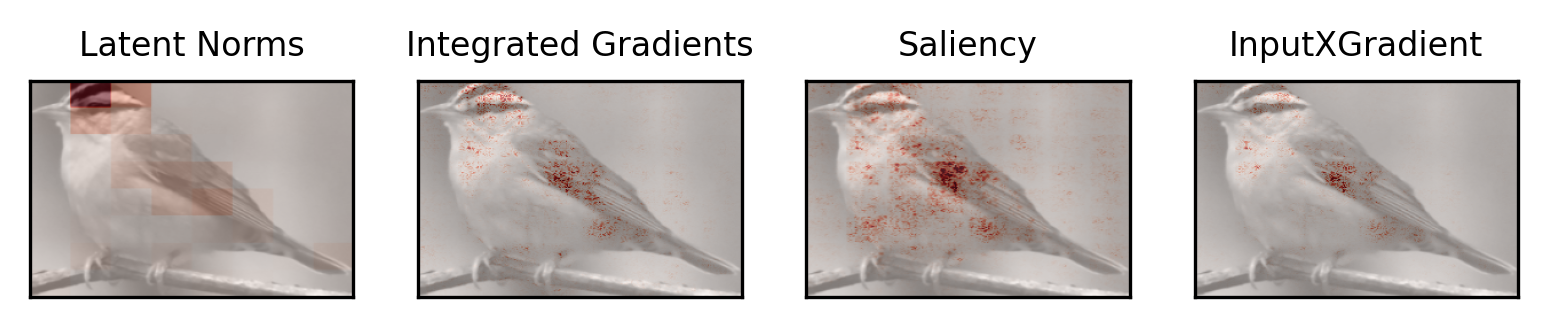}
         \caption{Feature attribution comparisons.}
         \label{fig:feat_attr_more_comparision_4}
     \end{subfigure}
     
     \begin{subfigure}[b]{\textwidth}
         \centering
         \includegraphics[width=\textwidth]{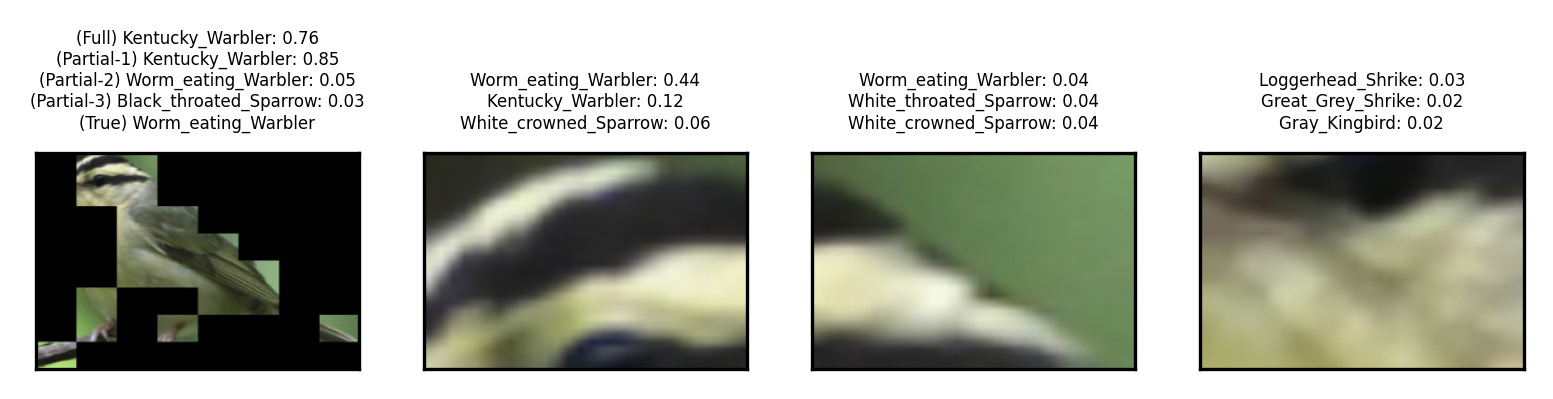}
         \caption{Algorithmic interpretation.}
         \label{fig:algo_more_comparision_4}
     \end{subfigure}

     \begin{subfigure}[b]{\textwidth}
         \centering
         \includegraphics[width=\textwidth]{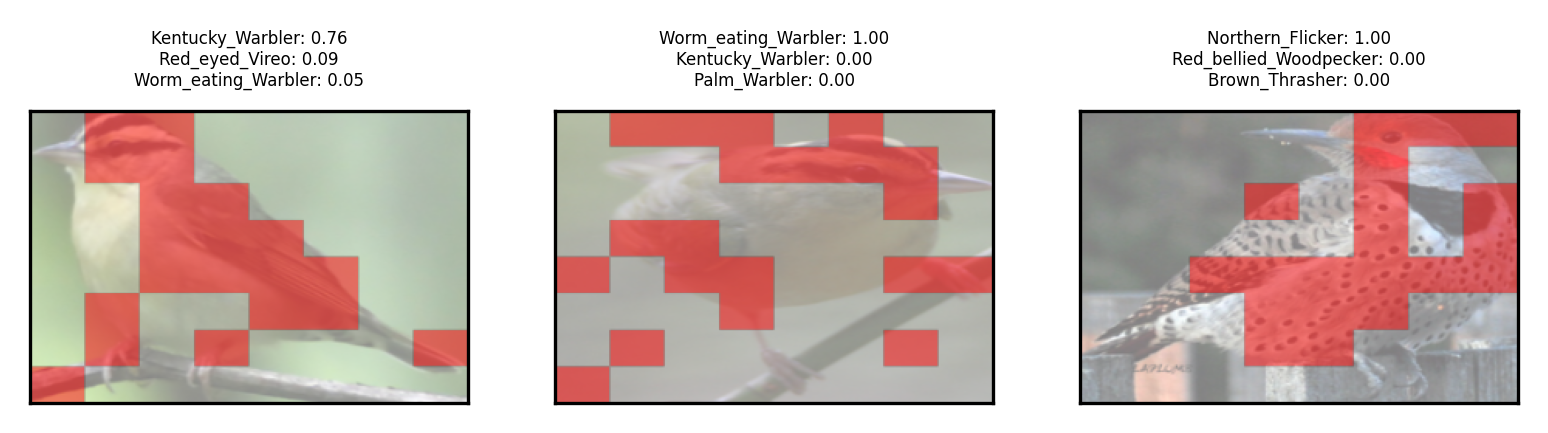}
         \caption{Example-based interpretation with correspondences.}
         \label{fig:example_more_comparision_4}
     \end{subfigure}
        \caption{Additional interpretability results.}
        \label{fig:app_cub_4}
\end{figure}

\clearpage

\subsubsection{TCR-EPITOPE Dataset}

In figure \ref{comppf}, we compare the interpretation of two examples by FLAN and pretrained FLAN. In FLAN the positions of the epitope sequence have considerably larger importance scores most likely due to the difference in the sequences' length, whereas pretrained FLAN manages to concentrate equally on both sequences. Furthermore, instead of considering the whole epitope sequence as important, pretrained FLAN focuses on specific positions and gives little weight to the rest.

\begin{figure}[h]
\centering
\caption{Interpretation of a binding and a non binding pair with pretrained FLAN.}
\label{comppf2}
\subfloat[Binding pair]{\label{fig:a}\includegraphics[width=\textwidth]{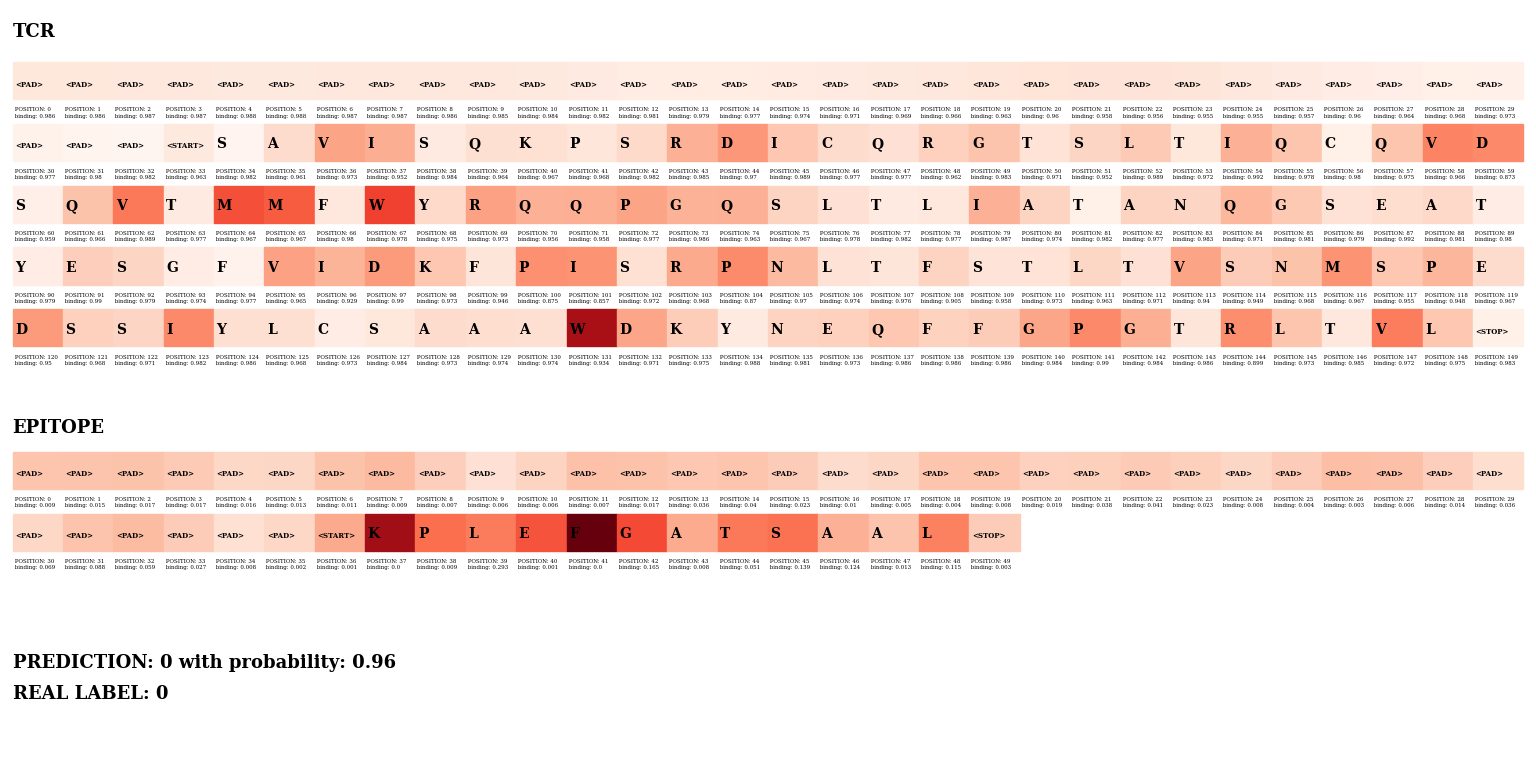}}\\
\subfloat[Non binding pair]{\label{fig:b}\includegraphics[width=\textwidth]{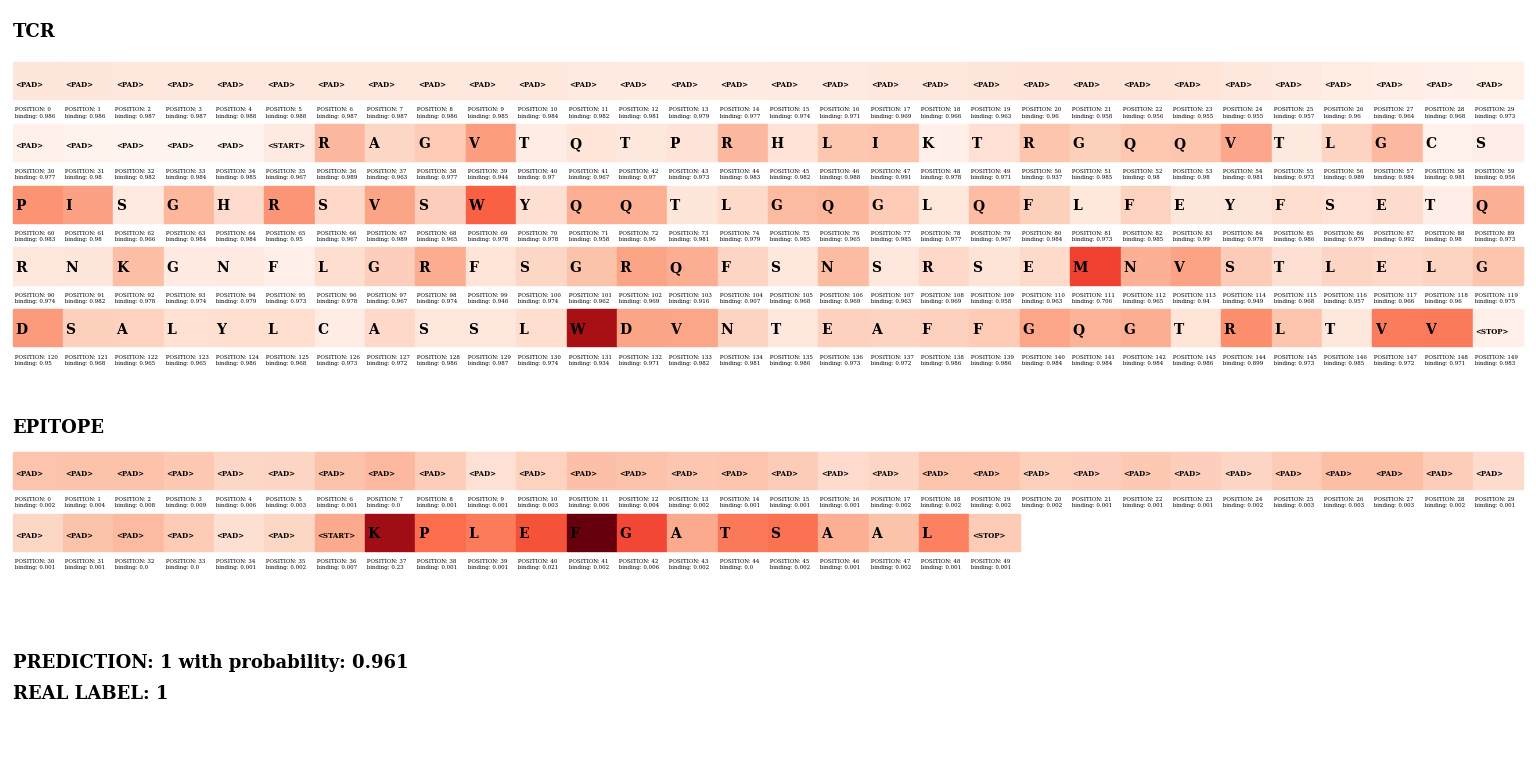}}%
\end{figure}

Figure \ref{comppf2} displays typical examples of interpreting a binding and a non-binding pair of sequences in pretrained FLAN. Both pairs have the same epitope and are classified correctly with high probability. Since the feature attributions are calculated independently for the two sequences, the epitope scores are the same in both examples. It is clear that pretrained FLAN focuses on specific positions in both sequences and in both examples. Moreover, the pretraining step suggested that certain amino acids are more important for the sequence representation, which may explain why they tend to have large scores independently of their position and the pair of sequences they belong to. For example, the "M" and "W" amino acids tend to have a very large weight in most examples and in most positions, a knowledge that is transferred to our model by the pretraining step.

Underneath each position there is its effect on the model's prediction. More specifically, let us assume we are interested in analyzing the effect of position 41 of the epitope sequence. We calculate a feature's effect by keeping all the positions of the TCR in the model and removing the remaining positions of the epitope sequence to get the binding probability predicted by the model. FLAN tends to concentrate on positions which seem irrelevant with the final prediction and also performs poorly with a linear classifier. Thus, there is evidence that the classifier is highly non-linear and the effect of each feature should not be approximated with this method.

\begin{figure}
    \caption{Comparison between FLAN and pretrained FLAN}
    \label{comppf}
    \centering 
    \subfloat[FLAN binding pair]{\label{fig:a}\includegraphics[width=\textwidth]{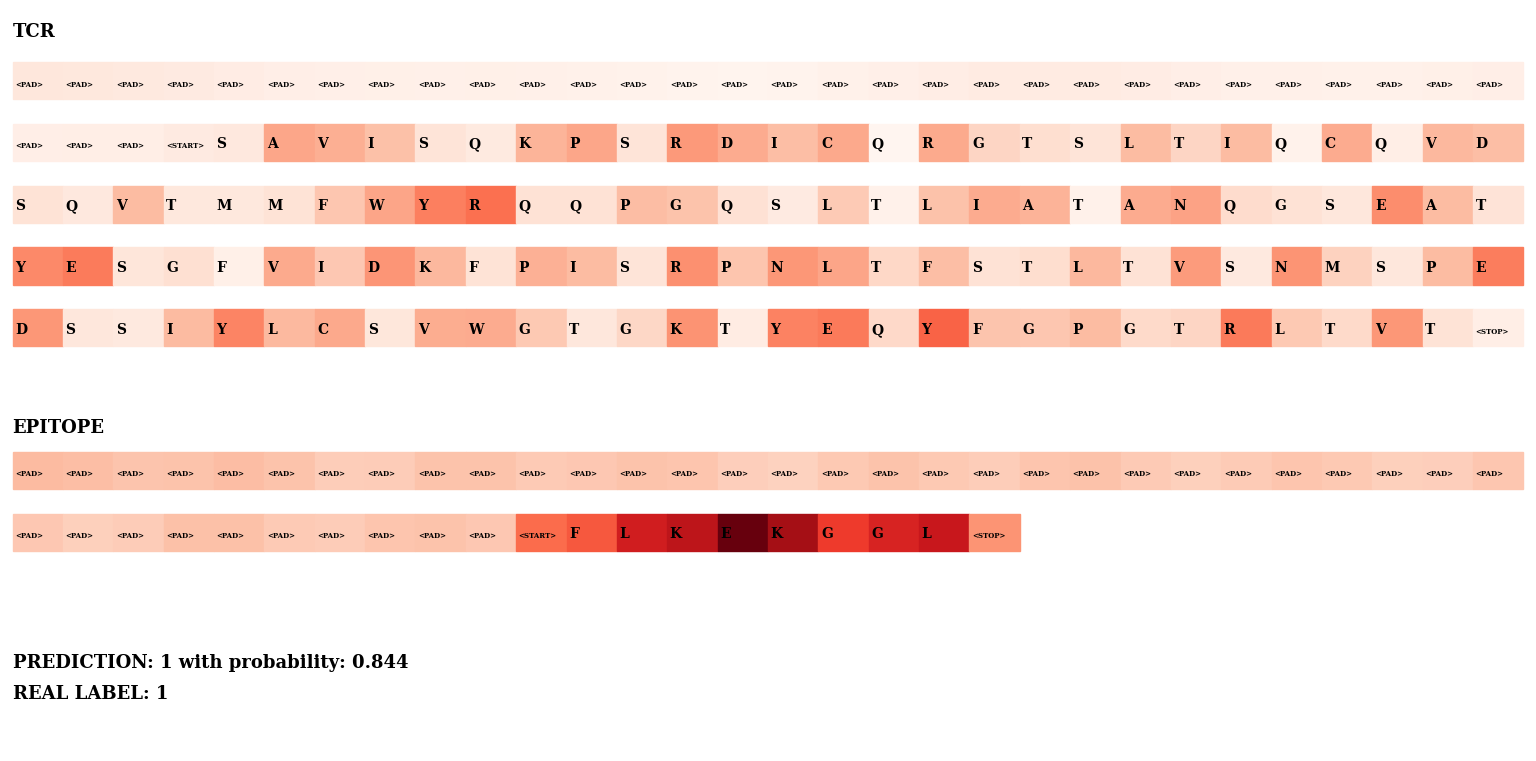}}\\
     \subfloat[FLAN non binding pair]{\label{fig:b}\includegraphics[width=\textwidth]{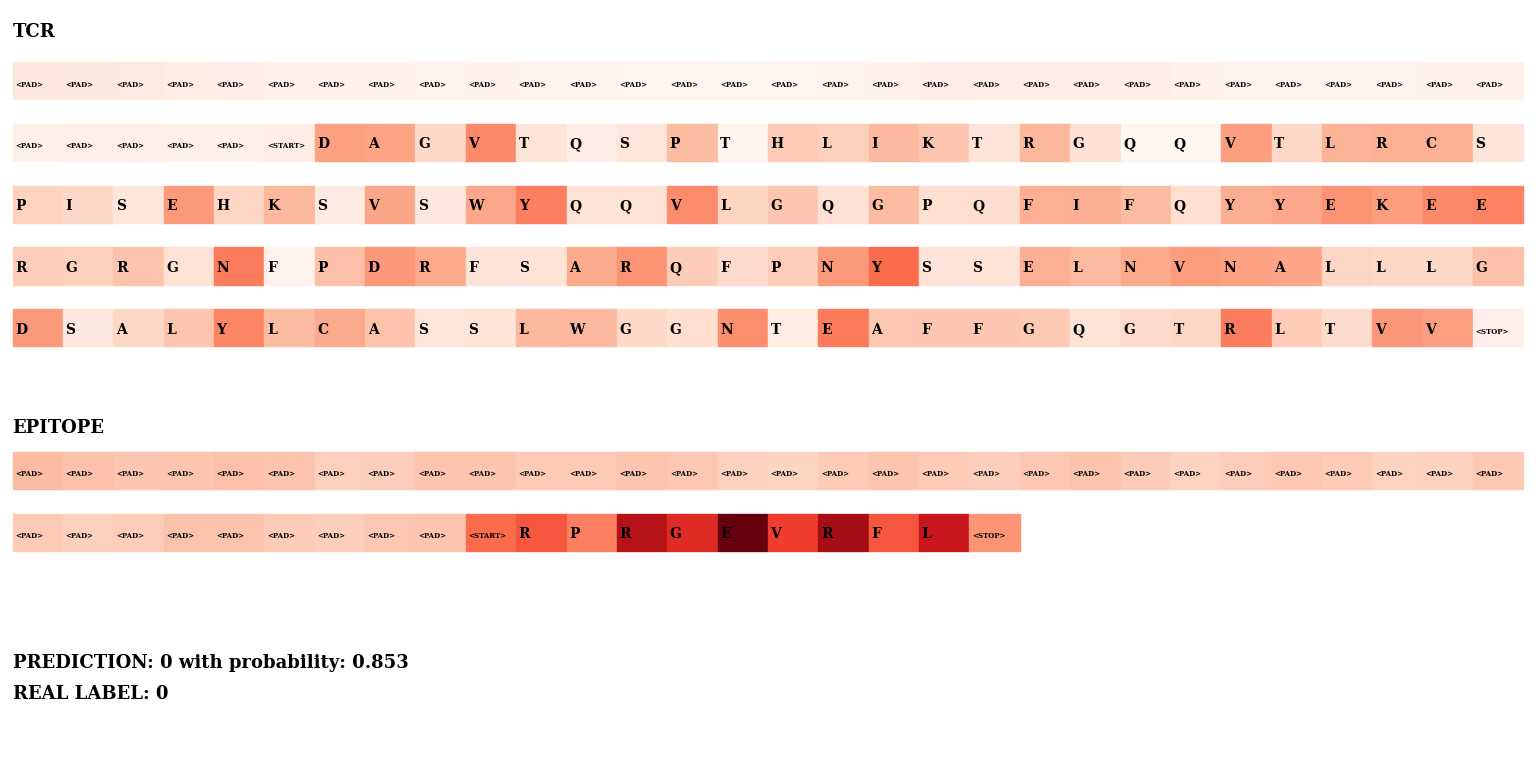}}\\
\end{figure}

\begin{figure}
  \ContinuedFloat 
  \centering 
  \subfloat[Pretrained FLAN binding pair]{\label{fig:c}\includegraphics[width=\textwidth]{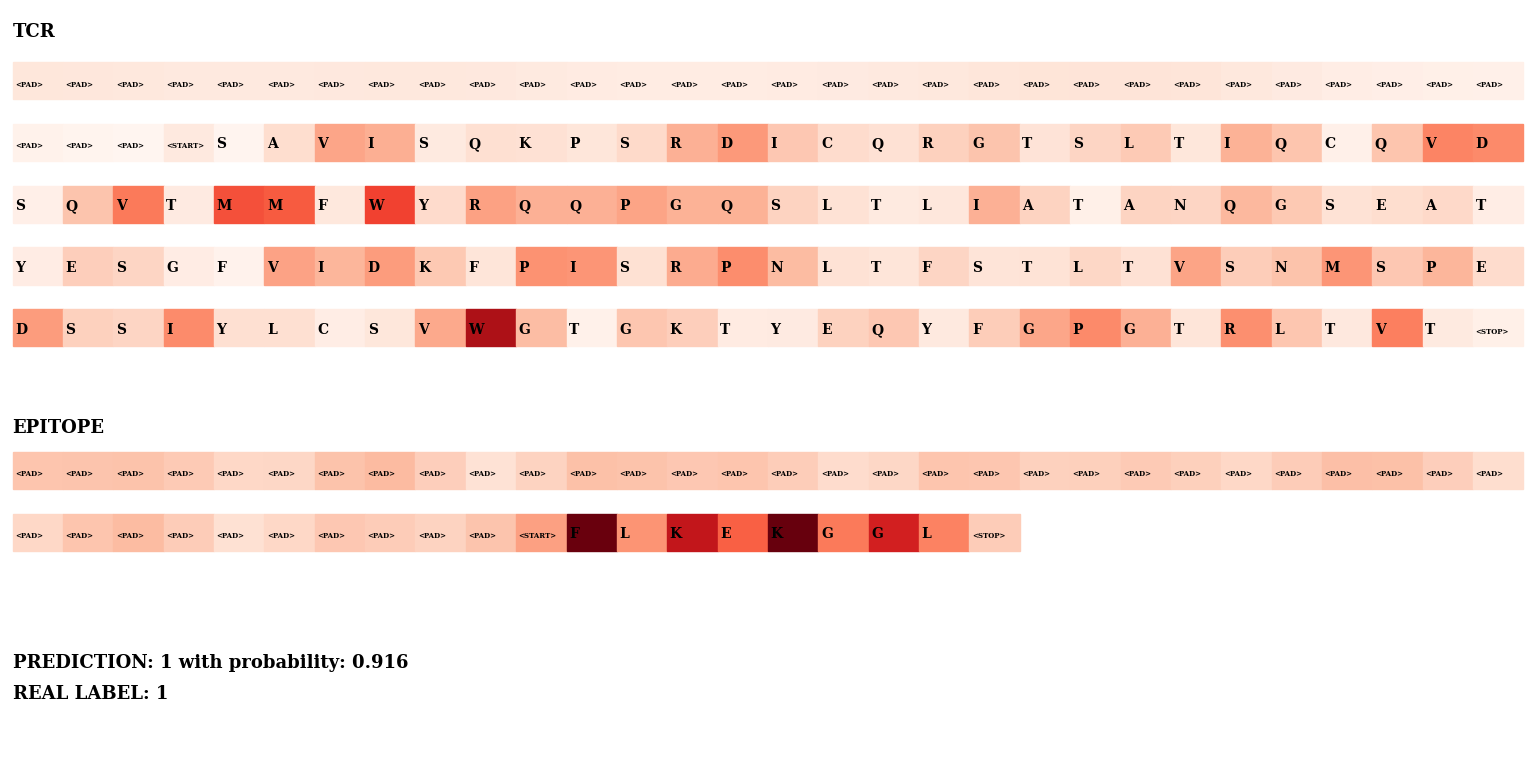}}\\
  \subfloat[FLAN non binding pair]{\label{fig:b}\includegraphics[width=\textwidth]{figures/Feat4105.png}}\\
\end{figure}

\clearpage

\subsection{Evaluation Metrics}
The tables for the example-based and feature attribution metrics with the corresponding standard deviations.

\begin{table}[h]
    \centering
    \caption{Example-Based Metrics for FLAN}\label{tab:ex_metrics_app}
    \resizebox{\linewidth}{!}{
    \begin{tabular}{lcccccccc}
        \toprule
        & \multicolumn{2}{c}{TCR-EPITOPE} & \multicolumn{2}{c}{MNIST} & \multicolumn{2}{c}{CUB}  & \multicolumn{2}{c}{SVHN}\\
    
    & OS & LS & OS & LS & OS & LS & OS & LS \\
         
    \midrule
    Local diversity & 299.267 $\pm$94.3 & 51.85 $\pm$7.5 & 68.088 $\pm$8.41 & 66.088 $\pm$18.099 & 1507.175 $\pm$270.52 & 1187.724 $\pm$102.77 & 67.239 $\pm$38.07 & 203.82 $\pm$58.94\\
    Local non-represent. & 2.21 $\pm$1.21 & 2.388 $\pm$1.69 & 2.154 $\pm$0.02 & 2.191 $\pm$0.05 & 3.633 $\pm$0.2 & 3.469 $\pm$0.31 & 2.139 $\pm$0.09 & 2.134 $\pm$0.08\\
    Global diversity & 671.458 $\pm$29.75 & 67.738 $\pm$0.9 & 61.012 $\pm$34.39 & 60.490 $\pm$33.47 & 1376.058 $\pm$3.81 & 1332.062 $\pm$463.64 & 185.147 $\pm$62.44 & 190.119 $\pm$85.8\\
    Global non-represent. & 0.178 $\pm$0.03 & 0.246 $\pm$0.04 & 2.162 $\pm$0.02 & 60.49 $\pm$33.47 & 3.81 $\pm$0.16 & 3.825 $\pm$0.16 & 2.091 $\pm$0.03 & 2.089 $\pm$0.03 \\
    \bottomrule
    * OS: Original Space , LS: Latent Space
  \end{tabular}
  }
  \end{table}

\vspace{1ex}

\begin{table}[h]
\caption{Feature Metrics}
\vspace{1ex}

    \begin{subtable}[h]{\linewidth}
    \centering
    \caption{Monotonicity}
    
    \resizebox{0.6\linewidth}{!}{
    \begin{tabular}{lcccc}
    \toprule
         & \texttt{TCR-EPITOPE}  & \texttt{MNIST} & \texttt{CUB} & \texttt{SVHN} \\
    \midrule
    FLAN & 0.173 $\pm$0.14 & 0.1 $\pm$0.08 & 0.024 $\pm$0.03 & 0.043 $\pm$0.03\\ 
    IntegratedGradients & 0.072 $\pm$0.05  & 0.15 $\pm$0.11 & 0.031 $\pm$0.02 & 0.044 $\pm$0.01\\
    InputXGradient & 0.071 $\pm$0.05 & 0.16 $\pm$0.1 & 0.044 $\pm$0.03 & 0.026 $\pm$0.02\\
    Saliency & 0.102 $\pm$0.06 & 0.15 $\pm$0.08 & 0.057 $\pm$0.06 & 0.035 $\pm$0.02\\
    \bottomrule
    \end{tabular}
    }
    
    \end{subtable}
 
    \vspace{1ex}
  
    \begin{subtable}[h]{\linewidth}
    \centering
    \caption{Non Sensitivity}
    
    \resizebox{0.6\linewidth}{!}{
    \begin{tabular}{lcccc}
    \toprule
         & \texttt{TCR-EPITOPE}  & \texttt{MNIST} & \texttt{CUB} & \texttt{SVHN} \\
    \midrule
    FLAN & 52.892 $\pm$26.39 & 483 $\pm$68.34 & 1886 $\pm$2574.75 & 27.285 $\pm$20.67\\ 
    IntegratedGradients & 25.191 $\pm$46.12 & 1.83 $\pm$0.98 & 5.4 $\pm$3.13 & 4.571 $\pm$3.95\\
    InputXGradient & 24.04 $\pm$46.43 &  3.5 $\pm$2.12 & 5.8 $\pm$2.58 & 5.142 $\pm$4.37\\
    Saliency & 23.502 $\pm$46.93 & 168.85 $\pm$20.32 & 111501.2 $\pm$14985.59 & 1015.44 $\pm$254.43\\
    \bottomrule
  \end{tabular}
  }
  
  \end{subtable}
  
\end{table}

\clearpage

\section{Supplemental files}

Config files and checkpoints of trained models are available at \url{https://drive.google.com/file/d/1uZ33GhdCKXECH9KjfGTUpCGTAfD9yEEe/view?usp=sharing}.
\end{document}